\documentclass[preprint,12pt]{elsarticle}

\biboptions{numbers,sort&compress}
\usepackage{amssymb}
\usepackage{amsmath}
\usepackage{subfig}
\usepackage{booktabs}
\usepackage{array}
 \usepackage{multirow}
\usepackage{algorithm} 
\usepackage{algpseudocode} 

\usepackage[
    colorlinks=true,    
    linkcolor=cyan,  
    urlcolor=cyan,   
    citecolor=cyan   
]{hyperref}
\usepackage{doi}

\newcommand{\figref}[1]{%
    \hyperref[#1]{%
      \textcolor{cyan}{Fig.}%
       \textcolor{cyan}{\ref*{#1}}%
    }
}

\renewcommand{\algref}[1]{%
    \hyperref[#1]{%
      \textcolor{cyan}{Algorithm}%
       \textcolor{cyan}{\ref*{#1}}%
    }
}

\newcommand{\tabref}[1]{%
    \hyperref[#1]{%
      \textcolor{cyan}{Table }%
       \textcolor{cyan}{\ref*{#1}}%
    }
}

\renewcommand{\eqref}[1]{%
  \hyperref[#1]{%
    \textcolor{cyan}{(}%
    \textcolor{cyan}{\ref*{#1}}%
    \textcolor{cyan})}%
 }

\journal{Neurocomputing}

\begin{document}

\begin{frontmatter}



\title{PoiCGAN: A Targeted Poisoning Based on Feature-Label Joint Perturbation in Federated Learning}


\author{Tao Liu} 
\ead{ltaoheu@163.com}
\author{Jiguang Lv\texorpdfstring{\corref{cor1}}{}}
\ead{lvjiguang@hrbeu.edu.com}
\author{Dapeng Man}

\author{Weiye Xi}

\author{Yaole Li}

\author{Feiyu Zhao}

\author{Kuiming Wang}

\author{Yingchao Bian}

\author{Chen Xu}

\author{Wu Yang}


\affiliation{organization={College of Computer Science and Technology},
            addressline={ Harbin Engineering University}, 
            city={Harbin},
            postcode={150001},
            country={China}}

\cortext[cor1]{Corresponding author}

\begin{abstract}
Federated Learning (FL), as a popular distributed learning paradigm, has shown outstanding performance in improving computational efficiency and protecting data privacy, and is widely applied in industrial image classification. However, due to its distributed nature, FL is vulnerable to threats from malicious clients, with poisoning attacks being a common threat. A major limitation of existing poisoning attack methods is their difficulty in bypassing model performance tests and defense mechanisms based on model anomaly detection. This often results in the detection and removal of poisoned models, which undermines their practical utility. To ensure both the performance of industrial image classification and attacks, we propose a targeted poisoning attack, PoiCGAN, based on feature-label collaborative perturbation. Our method modifies the inputs of the discriminator and generator in the Conditional Generative Adversarial Network (CGAN) to influence the training process, generating an ideal poison generator. This generator not only produces specific poisoned samples but also automatically performs label flipping. Experiments across various datasets show that our method achieves an attack success rate 83.97\% higher than baseline methods, with a less than 8.87\% reduction in the main task's accuracy. Moreover, the poisoned samples and malicious models exhibit high stealthiness.

\end{abstract}




\end{frontmatter}







\section{Introduction}
Industrial image classification~\cite{yang2024slsg,chen2024unified} plays an essential role in industrial vision applications~\cite{ettalibi2024ai,li2025survey}, covering tasks such as defect detection and surface analysis. Defect detection~\cite{peng2024industrial,chen2023surface} aims to identify faulty regions on a product’s surface or structure, such as cracks or scratches, while surface analysis~\cite{bai2025comprehensive} involves the detection and classification of attributes like texture and roughness on materials or objects' surfaces. The development of artificial intelligence technologies~\cite{escobar2021quality,gao2022review} has significantly enhanced the automation of these tasks, greatly improving overall production efficiency. However, traditional centralized paradigms face limitations due to the bottleneck of server computing capabilities and high data transmission costs~\cite{javaid2022exploring}. To address these challenges, a Federated Learning (FL)-based industrial image classification system~\cite{hegiste2024collaborative} has recently been proposed. FL~\cite{yurdem2024federated} is a distributed computing paradigm with privacy protection capabilities , enabling collaborative training by sharing model updates instead of traditional data-sharing methods.

Despite its advantages, the decentralized design of federated learning exposes the system to security risks posed by dishonest clients~\cite{feng2025survey}. Among these threats, poisoning attacks are particularly prevalent~\cite{almutairi2023federated,lu2022defense}. In this scenario, attackers may corrupt the learning process by introducing malicious training samples or by altering local model updates, which degrades local task performance and, after aggregation, undermines the effectiveness of the global model~\cite{moshawrab2024securing,wang2025enhancing}.

Most existing research on poisoning attacks primarily focuses on image classification tasks and can be categorized into data poisoning and model poisoning~\cite{wang2024fedtop}. In data poisoning~\cite{zhao2025data,yang2024invisible}, attackers often construct malicious samples by flipping labels or adding specific perturbations to the original images, thus degrading the model's performance through the training of these poisoned samples. In model poisoning~\cite{arazzi2025evading,khraisat2025securing}, attackers directly influence the global model's performance by modifying local model parameters or scaling malicious updates submitted to the server. However, a significant limitation of traditional poisoning attacks is the noticeable degradation in the performance of the main task, making them detectable through model performance monitoring. Another limitation is the substantial anomaly between the local models of malicious clients and benign users, which makes it difficult for these attacks to evade common detection mechanisms. These factors limit the practical impact of poisoning attacks in FL-based industrial image classification systems.

The significant decline in the global model's main task performance occurs because attacks lack a clear target, such as randomly flipping labels. This results in severe conflicts between the attack task and the main task, where the attacker sacrifices the main task's performance to enhance the attack's effectiveness. The significant anomaly in the malicious client’s model arises because it receives notably abnormal inputs during training. For instance, label-flipping attacks~\cite{lavaur2025investigating} corrupt the learning process by assigning incorrect annotations to training data, whereas image perturbation attacks distort input representations by introducing subtle modifications to sample features. When such compromised data are used for model optimization, the learned decision patterns can deviate substantially from their intended behavior. Therefore, attackers urgently need a targeted poisoning attack where the anomalies of the poisoned samples and models are less obvious.

To mitigate this issue, a targeted poisoning strategy termed PoiCGAN is introduced. The proposed approach leverages dual-feature collaborative perturbations to simultaneously preserve main-task accuracy while enhancing the stealthiness of malicious client models. Inspired by the concept of Conditional Generative Adversarial Networks (CGANs), we design a discriminator that not only distinguishes between real and fake images but also classifies images into specified categories. During the standard CGAN discriminator training process, we introduce samples where the image and label are misaligned as inputs. This adjustment to the discriminator's decision criteria guides the generator's training, enabling it to generate samples that do not align with the condition labels. To minimize the impact on the main task, we use one-to-one targeted attack setups, where a source class and a target class are defined. The goal of the attack is to misclassify images from the source class as belonging to the target class. Finally, we choose the target label as the conditional information for the CGAN. Following the above training process, the generator produces images from the source class that correspond to the target label, thus constructing poisoned samples through label-feature collaborative perturbation. Notably, we empirically control the number of CGAN training iterations within an optimal range to ensure the stealthiness of the malicious model.

The main contributions of this work can be summarized as follows:

\begin{itemize}
    \item We introduce PoiCGAN, a targeted poisoning framework designed for federated industrial image classification, which achieves high attack effectiveness while maintaining strong stealthiness. By leveraging targeted attacks and collaborative perturbations, our method preserves main task performance and reduces the significant anomaly of the malicious model. This work reveals a new vulnerability in such systems and lays the groundwork for more robust learning defenses.
    \item We propose the core module Poison Sample Generator (PSG), which generates poisoned samples by modifying CGAN discriminator training to induce automatic label flipping. Using a one-to-one attack setup and a target label as conditional information to achieve a targeted attack with less impact on the main task. Iteration count is tuned to control perturbation and maintain model stealth.
    \item Experiments conducted on multiple datasets demonstrate that poiCGAN achieves an average ASR improvement of 83.97\% over baseline methods, with less than a 8.87\% drop in main task accuracy. The evaluation also highlights its superior stealthiness of the malicious model and robustness against advanced defense mechanisms.
\end{itemize}

\section{Relate work}
In FL, poisoning threats can generally be divided into two categories: model-oriented attacks and data-oriented attacks, both of which are carried out by compromised clients~\cite{sagar2023poisoning}. The following sections provide a detailed discussion of each attack type and introduce some advanced defense methods.

\subsection{Model poisoning}
In model poisoning scenarios, malicious clients manipulate local update values during training, thereby undermining the integrity of individual models and causing adverse effects to spread to the global model after aggregation. Bhagoji et al.~\cite{bhagoji2019analyzing} first revealed the threat of model poisoning attacks in FL by demonstrating how a single malicious agent can cause misclassification in the global model by manipulating model updates. They proposed simple attack strategies, such as boosting malicious updates and alternating minimization. Yin et al.~\cite{yin2018byzantine} showed that distributed learning is particularly vulnerable to poisoning attacks due to the confidentiality of local data and models, and they implemented model poisoning via local model replacement techniques. To protect FL systems from these threats, researchers have developed numerous robust methods to defend against the aforementioned attacks. However, as attack techniques have evolved, a large number of model poisoning attacks targeting robust FL systems have emerged, which can easily bypass some existing defense mechanisms. Baruch et al.~\cite{baruch2019little} managed to bypass classic defenses such as Krum~\cite{blanchard2017machine} and Trimmed Mean~\cite{yin2018byzantine} with only minor but sophisticated modifications to model updates. Shejwalkar et al.~\cite{shejwalkar2022back} shared a similar viewpoint, suggesting that to evade defense mechanisms, attackers must ensure that the difference between malicious and benign updates is not too large. Building on Bhagoji’s work, Fang et al.~\cite{fang2020local} further explored the limitations of Byzantine-robust aggregation methods and designed a technique to directly modify local model parameters during training. This iterative process gradually deviates the global model from its original update direction. Zhou et al.~\cite{zhou2021deep} proposed an optimized model poisoning approach that exploits the invariance of redundant space in neural network training to inject malicious neurons, while simultaneously ensuring both attack performance and the main task's performance. To improve attack effectiveness, Sun et al.~\cite{sun2022semi} proposed a distance-aware model poisoning attack, which enhances attack strength by searching for the optimal target class in the feature space.

Despite significant advancements in these methods, they are more complex to implement, incur higher costs, and heavily rely on system permissions, making them more prone to detection and defense by the server compared to data poisoning attacks. As a result, their practicality is far inferior to that of data poisoning. The PoiCGAN we propose is a powerful data poisoning attack.

\subsection{Data poisoning}
Unlike model poisoning, data poisoning restricts adversaries to manipulating only a subset of local training data. The presence of such contaminated samples biases the learning process, leading to degraded local models whose negative effects are further amplified at the global level during aggregation. Based on whether label information is altered, data poisoning attacks are commonly divided into clean-label and label-flipping variants.

In clean-label attacks, the poisoned samples have correct labels, but their input features are carefully designed to cause the model to misclassify specific samples during testing. Attackers can generate poisoned samples using public datasets. Rong et al.~\cite{rong2022fedrecattack} simulated the feature distribution of local data based on public interaction and then trained the local model using this simulated distribution, causing it to degrade. However, this method assumes that client data are independently and identically distributed, which makes it difficult to deploy in the FL paradigm. Additionally, attackers can generate poisoned samples by adding perturbations to the training data, with one typical example being backdoor attacks~\cite{xie2019dba,liu2024beyond}. In backdoor attacks, attackers inject carefully designed triggers into clean samples and train the model on these samples to make incorrect predictions for specific inputs during inference. Recently, image perturbation methods based on Generative Adversarial Networks (GANs) have been widely used to generate poisoned data. To weaken the assumptions of data poisoning in FL, Zhang et al.~\cite{zhang2020poisongan} employed the global model as a discriminator throughout the training process to guide the generator in better learning the local data feature distribution and generating poisoned samples that resemble the original ones. To improve the trade-off between Attack Success Rate (ASR) and concealment, Sun et al.~\cite{sun2023vaguegan} reformulated the GAN optimization objective to alleviate excessive adversarial dynamics between the generator and discriminator. This modification enables the generator to embed malicious perturbations into synthesized samples, ultimately enhancing the overall attack capability.

Label-flipping attacks operate by corrupting the annotation process, causing training samples to be associated with incorrect class labels and thereby misleading the learned decision boundaries. As a consequence, models trained under such conditions tend to produce systematic misclassifications during inference. Based on the attacker’s intent, these attacks are commonly divided into targeted and non-targeted variants. In the former case, selected samples are reassigned to a specific adversarial class, whereas in the latter, labels are arbitrarily altered without a predefined target. Early studies in FL \cite{tolpegin2020data} demonstrated that targeted label manipulation could be exploited to generate poisoned datasets,  when incorporated into local training, introduce biased updates that propagate to the global model through aggregation. Although such strategies are straightforward to deploy, their attack effectiveness is often inconsistent and strongly influenced by the specific label-flipping scheme employed. To address this limitation, Gupta et al.~\cite{gupta2023novel} proposed a loss-based attack mechanism that modifies the optimization objective during training, steering the model away from its intended learning trajectory. By inducing erroneous associations between input samples and class labels, this approach achieves label-flipping effects in a more controlled and scalable manner.

However, these methods often struggle to balance attack strength and stealthiness. Clean-label attacks typically fail to achieve high ASR due to insufficient model learning, as the attack features are not distinct enough. Some image perturbation-based attacks can increase ASR by amplifying the perturbation, but this drastically alters the original features of the training samples, causing the malicious model to exhibit significant anomalies that are easy to detect by defense mechanisms. Label-flipping attacks, by directly modifying the labels, also result in obvious abnormal behavior in the malicious model. Therefore, we propose PoiCGAN, a feature-label collaborative perturbation-based attack. PoiCGAN reduces the behavioral discrepancy between malicious and benign models by carefully controlling the training duration during optimization.

\subsection{Defenses in FL}To mitigate the risks associated with poisoning attacks in FL paradigms, a range of defense mechanisms have been proposed. Cao et al.~\cite{cao2019understanding} detect abnormal client updates by measuring the pairwise Euclidean distances among local models and exclude suspicious contributions from the aggregation process. Cao et al.~\cite{shi2021federated} proposed a lightweight, unsupervised anomaly detection method based on support vector machines, which detects malicious models by examining the model's decision boundary. During model aggregation, only the benign local models are used to update the global model, thus mitigating the impact of malicious models on the global model. To ensure the inclusion of trustworthy participants, Cao et al.~\cite{cao2020fltrust} proposed a server-side defense strategy that leverages a trusted reference model trained on a clean dataset. By comparing this reference with client-submitted models, their approach enables the detection of anomalous updates and mitigates the influence of malicious contributions during aggregation. In addition, adjustments to the aggregation procedure further limit the impact of compromised clients on the global model. To cope with data heterogeneity across participants, Wang et al.~\cite{wang2020model} validated local models using a validation set, and the server only used the local models that performed well on this validation set to update the global model. Additionally, malicious models can also be identified by analyzing trends in model updates. Al Mallah et al.~\cite{al2023untargeted} used state persistence to monitor the training of all nodes. They assumed that attackers never truly trained their models but instead directly created model updates, identifying and removing malicious nodes by observing their behavior after a single iteration. Along this line, Zhang et al.~\cite{zhang2022fldetector} proposed a prediction-based defense that estimates expected model updates from each client’s historical update patterns and flags suspicious contributions when substantial discrepancies are observed between the estimated and reported updates.

However, these defense mechanisms rely on the assumption that the difference between malicious and benign models is sufficiently apparent. In Section \ref{5.3}, we assess the concealment of malicious models in PoiCGAN, and in Section \ref{5.5}, we validate the attack performance under three popular defense methods, demonstrating the robustness of the attack.

\section{Preliminaries}

\subsection{Federated learning}
FL~\cite{mcmahan2017communication} is a distributed learning paradigm with privacy-preserving characteristics. It aims to build a globally generalized model by collaboratively training participants' distributed datasets. The objective function is as follows:
\begin{equation}
\min _w f(w)=\sum_{i=1}^n p_i F_i(w)=E_i\left[F_i(w)\right]
\end{equation}
Here, $n$ refers to the number of participating clients in each training round, and $p_i$ indicates the selection probability of client $i$, subject to $p_i\geq0$ and $\sum_{i=1}^{n}P_{i}=1$. For client $i$, the local objective is defined as $F_{i}(w)=l(X_{i},Y_{i};w)$, which evaluates the empirical loss of model parameters $w$ on the corresponding data samples $(X_i,Y_i)$. The function $l(\cdot)$ specifies the task-dependent loss formulation.

A typical FL workflow involves three sequential phases. (1) Client selection: at the beginning of each training round, the central server samples a subset of available clients and distributes the current global model to the selected participants. (2) Local model update: each selected client performs local optimization using its private dataset to produce an updated local model, which is subsequently transmitted back to the server. (3) Model aggregation: the server integrates the received local updates to construct a new global model for the next training round. The aggregated global model is sent back to the designated clients for the next round of training. These three steps are repeated until the global model converges or the maximum number of training rounds is reached.

\subsection{Poisoning attacks in FL}
In FL, data poisoning attacks~\cite{sagar2023poisoning} manipulate client-side training data by perturbing input features or by corrupting label information, thereby influencing the learning process and degrading model performance. To achieve effective attacks, the attacker needs to carefully design the poisoned data to maximize the objective value in the following equation:

\begin{equation}
w^*=\arg \max _w \sum_{i=1}^n p_i F_i\left(w_i^t, \text{D}_{{poi}}\right)
\end{equation}
Here, $F_i(\cdot)$ represents the loss function of the $i$-th client, $\text{D}_{{poi}}$ refers to the poisoned dataset, $w_i^t$ denotes the local model parameters of client $i$ in the $t$-th round, and $p_i$ is the weighting factor.

The left half of \figref{fig1} provides an intuitive illustration of the data poisoning attack process in FL. Specifically, the attacker controls a subset of clients and introduces poisoned samples into their local training data. The construction of poisoned samples can be achieved by simply modifying the features or labels of clean samples~\cite{feng2025survey}, or by specially designing them. Using the crafted samples, adversaries generate biased local updates that are submitted to the server and incorporated during aggregation, which gradually degrades the integrity of the global model. Ultimately, the infected global model leads to incorrect predictions during the prediction phase or fails to converge.

\subsection{CGAN}
\label{3.3}
CGAN~\cite{mirza2014conditional} extend the standard GAN framework~\cite{goodfellow2014generative} by incorporating auxiliary condition variables $y$ into both the generator $G$ and the discriminator $D$. This conditional mechanism enables the generation process to be guided by external information, such as class annotations or complementary modality cues, thereby providing finer control over the characteristics of synthesized data. During training, the generator and discriminator are jointly optimized through an adversarial learning process, which can be formulated as the following objective function:

\begin{equation}
\begin{split}
\min_G\max_D V(D,G) = \mathbb{E}_{\boldsymbol{x}\sim p_{\text{data}}(\boldsymbol{x})}[\log D(\boldsymbol{x}\mid \boldsymbol{y})] 
\\  & \hspace{-11.5em}+ \mathbb{E}_{\boldsymbol{z}\sim p_z(\boldsymbol{z})}[\log (1-D(G(\boldsymbol{z}\mid \boldsymbol{y})))]
\end{split}
\end{equation}
In this formulation, $\log D(\boldsymbol{x}\mid \boldsymbol{y})$ denotes the discriminator’s estimated likelihood that sample $x$, conditioned on $y$, originates from the real data distribution, while $G(\boldsymbol{z}\mid \boldsymbol{y})$ represents the sample synthesized by the generator from noise $z$ under the same condition. The discriminator is updated to distinguish between authentic and generated samples while simultaneously verifying their consistency with the given condition, whereas the generator is trained to produce condition-consistent outputs that can effectively deceive the discriminator.

Under the conditional setting, the discriminator is optimized to distinguish real samples from generated ones while accounting for the given condition $y$. Its objective can be expressed as,

\begin{equation}
L_D=-\mathbb{E}_{x \sim p_{\text {data }}}[\log D(x \mid y)]-\mathbb{E}_{z \sim p_z}[\log (1-D(G(z \mid y)))]
\end{equation}

Conversely, the generator is trained to produce condition-consistent samples that maximize the discriminator’s confusion. The corresponding optimization objective for the generator is defined as,
\begin{equation}
\mathcal{L}_G=-E_{z \sim p_z(z)}[\log D(G(z \mid y))]
\end{equation}

The goal of the generator is to maximize the discriminator's error in classifying the generated samples (i.e., fake samples). In other words, the generator aims to produce fake samples that are likely to be classified as real by the discriminator, while also satisfying the conditional information $y$.

\section{Methodology}
This chapter begins by introducing the threat model employed by PoiCGAN, including a description of the attack scenarios, the attacker’s goals, knowledge, and capabilities. It then provides an overview of the workflow of PoiCGAN, followed by a detailed explanation of the core module, PSG.

\subsection{Threat model}
\subsubsection{Attack scenario}
We adopt the industrial image classification system proposed by He et al.~\cite{he2019semi} as the base framework for the main task and modify it into a distributed version based on FL for defect detection and surface analysis tasks. Following the attack configuration described in Tolpegin et al.~\cite{tolpegin2020data}, we assume that an adversary compromises a subset of participating clients and injects poisoned data during local training. These manipulated local updates impair individual model performance and, once aggregated, progressively deteriorate the quality of the global model. In addition, the attacker may amplify submitted updates prior to aggregation to further bias the global optimization process.

However, as mentioned earlier, aggressive performance degradation of the global model can easily expose malicious behavior and hinder attack success. Therefore, we adopt a one-to-one targeted attack~\cite{macas2024adversarial} setting that minimizes the impact on the main task performance as much as possible. This attack specifies only one source class and one target class, meaning that during the model training process, there exists only one type of poisoned sample that carries distinct features of the source class and the target label.

\subsubsection{Attacker goals}
Under the one-to-one targeted attack scenario, the adversary pursues two complementary objectives. The first objective focuses on attack effectiveness, where the global model $GM$ is manipulated such that the poisoned model $GM_{poi}$ consistently produces an incorrect target prediction $t$ for inputs belonging to a specific source class $s$, i.e., $ f(GM_{poi},x)=t\neq s=f(GM,x)$ for all $ x\in \text{D}_{s}$. Here, $\text{D}_{s}$ denotes the set of samples drawn from the source class.

The second objective emphasizes stealthiness. To avoid detection, the poisoned model is expected to preserve its original behavior on inputs outside the source set, such that $ f(GM_{poi},x)=f(GM,x)$ for all $ x\notin \text{D}_{s}$. These two requirements together ensure that the attack achieves targeted misclassification while maintaining high fidelity to the benign model’s behavior elsewhere, as follows:

\begin{equation}
f\left(GM_{poi}, x\right)=\left\{\begin{array}{cc}
t \neq s = f(GM, x) & \forall x \in \text{D}_s \\
f(GM, x) & \forall x \notin \text{D}_s
\end{array}\right.
\end{equation}

In addition, the attacker should minimize the visual differences between the poisoned sample and the original sample as much as possible. We validate this in Section \ref{5.2} through the visualization of poisoned and original samples. At the model level, stealthiness is also reflected in the similarity between the poisoned and benign parameter distributions. To assess this, we introduce a metric called the Model Indistinguishability Score (MIS) in Section \ref{5.1.4}, which evaluates the stealthiness of the poisoned model in the parameter space, and present the evaluation results in Section \ref{5.3}.

\subsubsection{Attacker knowledge and capabilities}
\label{4.1.3}
Following the Kerckhoffs’s principle~\cite{shannon1949communication}, we adopt an adversary model consistent with prior federated learning security studies, such as Tolpegin et al.~\cite{tolpegin2020data}. In this setting, the adversary is assumed to compromise $k$ participating clients and gain access to their local training data, optimization procedures, and model parameters. The proportion of compromised clients is quantified by the Poisoned Model Rate (PMR), defined as $\text{PMR} = k/N$, where $N$ denotes the total number of clients.

In addition, the adversary is aware of the aggregation mechanism employed by the server, including any potential defense strategies applied during aggregation. Nevertheless, the attacker is restricted from influencing computations performed at the server side or interfering with the training processes of honest clients.

\subsection{Overall workflow}
Based on the above threat model, we now detail the proposed attack method. The left half of \figref{fig1} provides an intuitive overview of the overall workflow of PoiCGAN, with the specific execution steps outlined as follows:

\textbf{Step 1: Poisoned sample generation.} After identifying the controlled clients, the adversary uses a small number of original samples to generate poisoned samples that meet the required conditions, \textbf{utilizing the core module PSG}. These poisoned samples should carry distinct features of the source class image and the target label.

\textbf{Step 2: Local model training.} The poisoned samples generated in the previous step guide the local training of the malicious client's model, while benign clients use clean samples to train their local models.

\textbf{Step 3: Infection of the global model.} Following local optimization, the server aggregates a subset of client-submitted models to update the global model. In each training round, a fraction of participating clients is assumed to be compromised, with the proportion governed by the PMR. The global model obtained after aggregation will also be poisoned. Inspired by the model replacement concept proposed by Bag-
dasaryan et al.~\cite{bagdasaryan2020backdoor}, we further enhance the attack's performance by scaling the model parameters.

\textbf{Step 4: Model inference.} The final global model is capable of executing the targeted attack task while maintaining the performance of the main task. Specifically, it should correctly classify all test samples except those from the source class and predict the source class images as the target class.

Notably, the core contribution of our work lies in the PSG module used in the first step. We achieve targeted optimization of the CGAN training process by precisely controlling the input conditional information. By establishing a link between source class images and target labels, we mislead the discriminator into making incorrect predictions, thereby guiding the generator to automatically flip labels during the image generation process when given specified conditions as input.

\subsection{PSG module}
This section focuses on the composition and principles of the core module, PSG. Building upon prior studies on GAN-based data generation~\cite{zhang2020poisongan}, PSG leverages a generative adversarial framework to synthesize poisoned samples. As described earlier, we aim to generate poisoned samples that satisfy two conditions: (1) the targeted generation of poisoned images, meaning only images containing features of the source class should be generated; and (2) the automatic label flipping of the generated poisoned images, flipping them to the target label.

\begin{figure}[h]
\centering
\includegraphics[width=1\textwidth]{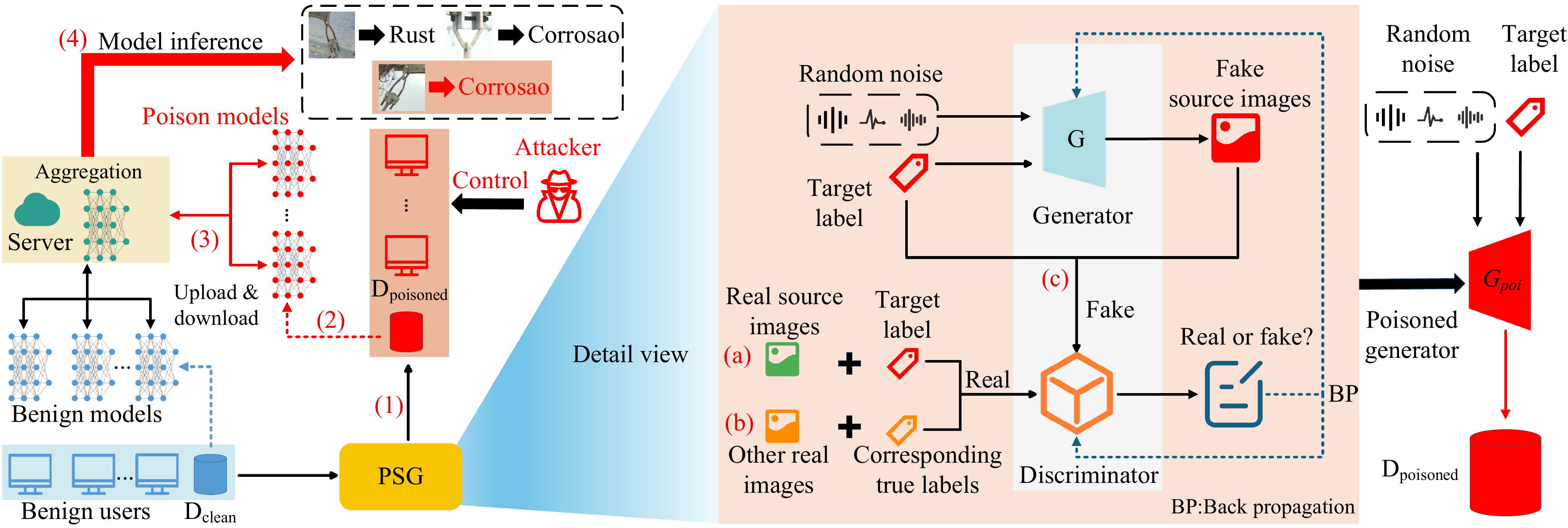}
\caption{Workflow of PoiCGAN and core module PSG. The left side of the figure shows PoiCGAN's workflow, comprising four steps: (1) poisoned sample generation, (2) local model training, (3) infection of the global model, and (4) model inference. In step (1), PSG generates poisoned samples by incorporating the target label as conditional input into both the discriminator and generator, as shown on the right side. The discriminator training includes: (a) ``real source images + target label" to output real, (b) ``real images from other classes" to output real, and (c) ``fake images + target label" to output fake. The goal of (a) is to mislead the discriminator, guiding the generator to flip the label and produce source class images when conditioned on the target label.}
\label{fig1}
\end{figure}

Considering that in Section \ref{3.3}, CGAN~\cite{ribas2025conditional} controls the behavior of the generator and discriminator by adding conditional information $c$, the competition between the two results in the generation of fake data by $G$ that is indistinguishable from the original data while meeting the condition $c$. Based on this idea, we propose the design of \textbf{a variant of the CGAN framework (PSG) to train an ideal poisoned image generator}, $G_{poi}$, which generates poisoned samples satisfying the above conditions. The key idea is to \textbf{use the target label as conditional information} to control the generator's targeted generation and to cause the discriminator to make incorrect predictions on source class images. Ultimately, $G_{poi}$ can generate images similar to source class images based on the input target label.

The detailed view of PSG is shown in the right half of \figref{fig1}, where the module is composed of a generator $G$ and a discriminator $D$. The generator synthesizes poisoned samples by taking random noise together with the target label as conditional input, with the objective of producing images that exhibit visual characteristics of the source class. The discriminator is trained using three types of inputs: receiving (1) ``source class real images + target label” which should be classified as real; (2) ``non-source class real images + corresponding real labels” which should also be classified as real; and (3) ``generated fake samples + target label” which should be classified as fake. The latter two categories follow the conventional CGAN training paradigm, where the discriminator is optimized to differentiate synthesized data from authentic samples. The key modification lies in the first part, where targeted label-flipping sample pairs are used to train the discriminator, enabling it to produce a specific classification criterion. The resulting feedback is then provided to the generator to guide its generation process. The specific execution steps of PSG are as follows:

\textbf{Step 1: Sampling real samples and corresponding labels.} A small number of real samples $x_{real}$ are sampled from the malicious client’s local training set, with each sample corresponding to a real label $y_{real}$. Then, the core operation is performed: the labels of all source class samples $x_s$ are changed from the source class label $s$ to the target label $t$, while the labels of other class samples $x_i (i \neq s)$ remain unchanged. This results in two types of data pairs: non-source class data pairs $(x_i, y_i)$ and source class data pairs with flipped labels $(x_s, t)$.

\textbf{Step 2: Sampling random noise and generating fake samples.} Random latent vectors $z$ are drawn from a predefined noise distribution, which is instantiated as a standard normal distribution $N(0,1)$ in our implementation. Together with the designated target label $t$, these latent variables condition the generator to synthesize fake samples, expressed as $x_{fake} = G(z|t)$.

\textbf{Step 3: Updating the discriminator.} The discriminator receives data pairs $(x, y)$ as input. Based on the previous two steps, non-source class data pairs $(x_i, y_i)$ and source class data pairs with flipped labels $(x_s, t)$ are used as positive samples, while the fake data pairs $(x_{fake}, t)$ are used as negative samples for training. The discriminator's loss function is as follows:

\begin{equation}
\begin{split}
\mathcal{L}_D=-\mathbb{E}_{x_i \sim p_{\text {data }}}[\log D(x_i \mid y_i)]-\mathbb{E}_{x_s \sim p_{\text {data }}}[\log D(x_s \mid t)] \\
& \hspace{-21.8em} -\mathbb{E}_{z \sim p_z}[\log (1-D(G(z \mid t) \mid t))], i\neq s
\end{split}
\end{equation}
Here, the loss component $-\mathbb{E}_{x_s \sim p_{\text{data}}}[\log D(x_s \mid t)]$ encourages the discriminator to associate source-class samples with the designated target label during training. Through back propagation, this guides the generator's learning process, enabling it to automatically flip labels during the fake sample generation process when the target label is provided as conditional information. We use gradient descent to optimize the discriminator's parameters, enabling it to better distinguish between real and fake samples.

\begin{algorithm}[h!]
\caption{PSG}
\label{alg1}
\begin{algorithmic}[1]
\State \textbf{Input:} malicious client local training set $\text{D}_m$, generator $G$, discriminator $D$, batch size $b$, number of iterations $T$, distribution of random noise $p_z$, target label $t$, source label $s$
\State \textbf{Output:} poison generator $G_{poi}$
\State Initialize the  $G$ and  $D$
\For{$n = 1, 2, \dots, T$}
    \State /* \textit{Sample real data and flip source class labels.}*/
    \State $(x_s,s) = \text{sample\_source\_images} \hspace{0.1em}(\text{D}_m,b)$
    \State $(x_s,t) \gets (x_s,s)$ //\textit{Flip the labels from $s$ to $t$.}
    \State $(x_i, y_i) = \text{sample\_non-source\_images} \hspace{0.1em} (\text{D}_m,b)$ 
    
    \State /* \textit{Sample noise and generate fake data.}*/
    \State $z = \text{sample\_noise}\hspace{0.1em}(p_z, b)$
    \State $x_{\text{fake}} = G(z,t)$
    
    \State /* \textit{Update discriminator.}*/
    \State $D_{\text{real\_non-s}} = D(x_i, y_i)$
    \State $D_{\text{real\_s}} = D(x_s, t)$
    \State $D_{\text{fake}} = D(x_{\text{fake}}.\text{detach()}, t)$
    \State $\mathcal{L}_D = - \left( \log(D_{\text{real\_non-s}}) + \log(D_{\text{real\_s}}) + \log(1 - D_{\text{fake}}) \right). \text{mean()}$
    \State $D.\text{optimizer.zero\_grad()}$
    \State $\mathcal{L}_D$.backward()
    \State $D.\text{optimizer.step()}$
    
    \State /* \textit{Update generator.}*/
    \State $z = \text{sample\_noise}\hspace{0.1em}(p_z, b)$ //\textit{Resample noise.}
    \State $x_{\text{fake}} = G(z,t)$
    \State $D_{\text{fake}} = D(x_{\text{fake}}, y)$
    \State $\mathcal{L}_G = - \left( \log(1 - D_{\text{fake}}) \right). \text{mean()}$
    \State $G.\text{optimizer.zero\_grad()}$
    \State $\mathcal{L}_G$.backward()
    \State $G.\text{optimizer.step()}$
\EndFor
\State $G_{poi} \gets G$
\State \textbf{return} $G_{poi}$
\end{algorithmic}
\end{algorithm}

\textbf{Step 4: Updating the generator.} One objective of the generator is to ``deceive the discriminator” i.e., to make $D(G(z|y)|y)$ approach 1. Another goal is to generate only poisoned samples that satisfy the condition, which is achieved by ensuring $y = t$. Guided by the discriminator’s feedback, the generator progressively learns to produce images that preserve source-class characteristics while being associated with the target label, thereby enabling automatic label flipping. The generator’s loss function is as follows:

\begin{equation}
\mathcal{L}_G=-\mathbb{E}_{z \sim p_z}[\log (1-D(G(z \mid t) \mid t))]
\end{equation}
Here, gradient descent is also used to optimize the generator's parameters.

\textbf{Step 5: Repeat the above steps several times} until the desired poisoned generator $G_{poi}$ is obtained. 

The process outlined above is detailed in Algorithm \ref{alg1}. A key advantage of this algorithm is that it requires only a small number of clean samples as input to train a powerful poison sample generator, $G_{poi}$, for generating the attacker’s desired poison samples.

\section{Experiments \& analyses}

\subsection{Experimental settings}

\subsubsection{General setups}
The experiments were conducted on a server equipped with two NVIDIA 3090 GPUs, using PyTorch~\cite{yuan2024performance} as the software framework. The performance of PoiCGAN was evaluated on three datasets and three models, as shown in \tabref{tab1}. In this context, Clean ACC denotes the classification accuracy of the global model on the main task after convergence under benign training conditions, i.e., in the absence of adversarial interference.

\begin{table}[h]
    \renewcommand{\arraystretch}{0.5} 
    \centering 
    \footnotesize 
    \caption{Datasets and models.}
    \label{tab1}
    \setlength{\tabcolsep}{1mm}{
        \setlength{\heavyrulewidth}{0.1mm}  
        \setlength{\lightrulewidth}{0.1mm}  
        \begin{tabular}{c c c c c c c }
        \toprule 
        \multicolumn{1}{c}{Datasets} &\multicolumn{1}{c}{Labels} &\multicolumn{1}{c}{Size} &\multicolumn{1}{c}{Training} &\multicolumn{1}{c}{Testing} &\multicolumn{1}{c}{Models} &\multicolumn{1}{c}{Clean ACC}\\[1.5pt]
        \midrule
        \multicolumn{1}{c}{InsPLAD-fault} &4 &128*128*3 &2250 &146 &ResNet-18 &88.4\% \\[1.5pt]
        \multicolumn{1}{c}{NEU-CLS} &6 &128*128*1 &1260 &540 &LeNet-5 &94.4\% \\[1.5pt]              
        \multicolumn{1}{c}{Kylberg} &6 &128*128*1 &3920 &560 &ResNet-34 &99.2\% \\[1.5pt]
        \bottomrule 
        \end{tabular}
    }           
\end{table}

The detailed description of each data set is as follows:

\textbf{NEU-CLS}~\cite{bouguettaya2023deep}: NEU-CLS is a public steel surface defect dataset released by Northeastern University. It consists of six defect categories--RS, Pa, Cr, PS, In, and Sc--with 300 grayscale images provided for each class. For experimental evaluation, all images were resized to a resolution of 128*128 pixels, and the dataset was partitioned into training and testing subsets following a 7:3 split.

\textbf{InsPLAD}~\cite{vieira2023insplad}: The InsPLAD-fault (referred to as InsPLAD) is a power line defect classification dataset released in 2023. It covers five types of assets and four defect types (rust, corrosion, missing-cap, bird-nest). The data is processed as 128*128 RGB images, and the training and testing data split is detailed in \tabref{tab2}.

\begin{table}[h]
    \centering
    \footnotesize 
    \renewcommand{\arraystretch}{1} 
    \caption{Data splitting details on InsPLAD-fault.}
    \label{tab2}
    \setlength{\tabcolsep}{0.65mm} 
    \setlength{\heavyrulewidth}{0.1mm}  
    \setlength{\lightrulewidth}{0.1mm}  
    \begin{tabular}{c @{\hspace{6mm}} c @{\hspace{6mm}}c }
    \toprule 
    Defect types &Training samples &Testing samples \\[1.5pt]
    \midrule
    Missing-cap &270 &30 \\[1.5pt]
    Corrosion &740 &33 \\[1.5pt]
    Bird-nest &350 &20 \\[1.5pt]
    Rust (three types of assets) &310/290/290 &20/23/20 \\[1.5pt]
    \bottomrule 
    \end{tabular}
\end{table}

\textbf{Kylberg texture dataset}~\cite{goyal2023texture}: Created by the University of Gothenburg, Sweden, this dataset includes 28 types of material textures (such as wood, stone, fabric, etc.), with 160 images per class. The dataset is widely used for texture classification and material recognition, which belong to the category of surface analysis tasks.

The models used for training are described as follows:

\textbf{LeNet-5}~\cite{an2024lightweight}: This network consists of two convolutional layers and three fully connected layers. Each convolutional layer is followed by batch normalization, ReLU activation, and a pooling layer, enabling effective feature extraction. The Adam optimizer is used with a learning rate of 0.0001 and a batch size of 8.

\textbf{ResNet-18}~\cite{nisa2025improved}: The network adopts a residual learning architecture composed of 17 convolutional layers followed by a single fully connected layer. By introducing shortcut connections across layers, the network effectively alleviates the vanishing gradient issue during training. A four-neuron fully connected layer with a Softmax activation is appended to produce the final classification outputs. The training parameters are similar to those of LeNet-5, but with a batch size of 64.

\textbf{ResNet-34}~\cite{shahin2025fine}: Similar to ResNet-18 in structure but deeper, with 33 convolutional layers, making it suitable for extracting more complex image features. The residual connection mechanism is also used to enhance model expressiveness and stability.

\subsubsection{Baseline setups}
This experiment compares and evaluates PoiCGAN's robust attack performance by using three advanced poisoning attacks in FL as baseline methods: Targeted Data Poisoning (TDP), Targeted Model Poisoning (TMP), and Attacking-Distance-Aware (ADA) Attacks. The following provides a detailed description of each baseline method. Unless otherwise specified, all baseline methods are implemented in accordance with the configurations reported in their original studies.

\textbf{TDP}~\cite{tolpegin2020data}: Tolpegin et al. conducted the first systematic analysis of malicious clients in FL, who manipulate local data or model updates to perform label flipping or parameter poisoning, thereby impairing the global model’s classification performance. TDP consists of two attack modes: data poisoning and training-phase attacks. In this study, we focus on the former, considering it as a data poisoning method for comparison with PoiCGAN.

\textbf{TMP}~\cite{bhagoji2019analyzing}: Bhagoji et al. revealed vulnerabilities in the FL aggregation process, enabling attackers to alternate between malicious and benign updates using a minimization strategy that mimics legitimate user behavior, thereby evading detection. TMP also introduces parameter evaluation techniques to enhance the attack's stealthiness.

\textbf{ADA}~\cite{sun2023attacking}: Sun et al. proposed a semi-targeted attack framework for scenarios with limited prior knowledge. The method leverages gradient information from the final network layer to facilitate target class selection, enabling both low-frequency and high-efficiency poisoning effects. The main feature of ADA is its use of inter-class attack distance quantification to guide optimal target selection.

\subsubsection{FL \& PSG setups}
\textbf{FL setups:} This experiment involves multiple participants and malicious clients. Considering the limited sample size, we assume a total of 20 clients. In each communication round, the server samples 10 clients to participate in model aggregation in order to promote efficient convergence of the global model. Aggregation is performed using the widely adopted FedAvg algorithm~\cite{mcmahan2016federated}. The participants consist of benign and malicious clients, with the ratio based on the PMR mentioned in Section \ref{4.1.3}. Here, we take PMR=40\% as the base setting.

To guarantee a sufficiently strong influence on the global model, two complementary attack strategies are adopted. First, we poison all local data~\cite{kumar2025minimal}, meaning that in the poisoned rounds, malicious clients train exclusively on poisoned samples. Second, we adopt a continuous poisoning setup, maintaining the same PMR and Poisoned Data Ratio from round 50 onwards. Local training is performed for 2 iterations per round, with a total of 200 rounds of FL training, and the scaling factor for aggregation is set to 1 by default.

\textbf{PSG setups:} To achieve high-quality poisoned sample generation, the generator and discriminator of the PSG module are implemented using a convolutional neural network architecture, following the design principles introduced by He et al.~\cite{he2019semi}.  The detailed network configuration is summarized in \tabref{tab3}. Unless otherwise specified, the number of training rounds for the PSG module is fixed at $T=200$ throughout the experiments.

\begin{table}[h]
    \renewcommand{\arraystretch}{0.5} 
    \centering 
    \footnotesize 
    \caption{Network architecture of the generator and discriminator.}
    \label{tab3}
    \setlength{\tabcolsep}{1mm}{
        \setlength{\heavyrulewidth}{0.1mm}  
        \setlength{\lightrulewidth}{0.1mm}  
        \begin{tabular}{c c c c @{\hspace{6mm}} c c c c}
        \toprule 
        \multicolumn{4}{c}{\textbf{Generator}} &\multicolumn{4}{c}{\textbf{Discriminator}} \\[1.5pt]
        \multicolumn{1}{c}{Name} &\multicolumn{1}{c}{Type} &\multicolumn{1}{c}{Filters} &\multicolumn{1}{c}{Size/stride} & \multicolumn{1}{c}{Name} &\multicolumn{1}{c}{Type} &\multicolumn{1}{c}{Filters} &\multicolumn{1}{c}{Size/stride} \\[1.5pt]
        \midrule
        \multicolumn{1}{c}{D1} &\multicolumn{1}{c}{deconv} &512 &4*4/1 &\multicolumn{1}{c}{C1} &\multicolumn{1}{c}{conv} &128 &4*4/2 \\[1.5pt]
        \multicolumn{1}{c}{B1} &\multicolumn{1}{c}{BN} &- &- &\multicolumn{1}{c}{B1} &\multicolumn{1}{c}{BN} &- &- \\[1.5pt]
        \multicolumn{1}{c}{D2} &\multicolumn{1}{c}{deconv} &256 &4*4/2 &\multicolumn{1}{c}{C2} &\multicolumn{1}{c}{conv} &128 &4*4/2 \\[1.5pt]
        \multicolumn{1}{c}{B2} &\multicolumn{1}{c}{BN} &- &- &\multicolumn{1}{c}{B2} &\multicolumn{1}{c}{BN} &- &- \\[1.5pt]
        \multicolumn{1}{c}{D3} &\multicolumn{1}{c}{deconv} &256 &4*4/2 &\multicolumn{1}{c}{C3} &\multicolumn{1}{c}{conv} &256 &4*4/2 \\[1.5pt]
        \multicolumn{1}{c}{B3} &\multicolumn{1}{c}{BN} &- &- &\multicolumn{1}{c}{B3} &\multicolumn{1}{c}{BN} &- &- \\[1.5pt]
        \multicolumn{1}{c}{D4} &\multicolumn{1}{c}{deconv} &128 &4*4/2 &\multicolumn{1}{c}{C4} &\multicolumn{1}{c}{conv} &256 &4*4/2 \\[1.5pt]
        \multicolumn{1}{c}{B4} &\multicolumn{1}{c}{BN} &- &- &\multicolumn{1}{c}{B4} &\multicolumn{1}{c}{BN} &- &- \\[1.5pt]
        \multicolumn{1}{c}{D5} &\multicolumn{1}{c}{deconv} &128 &4*4/2 &\multicolumn{1}{c}{C5} &\multicolumn{1}{c}{conv} &512 &4*4/2 \\[1.5pt]
        \multicolumn{1}{c}{B5} &\multicolumn{1}{c}{BN} &- &- &\multicolumn{1}{c}{B5} &\multicolumn{1}{c}{BN} &- &- \\[1.5pt] 
        \multicolumn{1}{c}{D6} &\multicolumn{1}{c}{deconv} &1/3 &4*4/2 &\multicolumn{1}{c}{C6} &\multicolumn{1}{c}{conv} &1 &4*4/1 \\[1.5pt]
        \bottomrule 
        \end{tabular}
    }
\end{table}

\subsubsection{Evaluation metrics}
\label{5.1.4}
\textbf{ACC:} The accuracy of the model in the main task, representing the proportion of test samples for which the global model provides correct predictions. It is calculated as follows:
\begin{equation}
\text{ACC}=N_{right}/N_{total}
\end{equation}
where $N_{right}$ is the number of correctly classified samples and $N_{total}$ is the total number of test samples. In PoiCGAN, the attacker's goal is to minimize the impact on ACC and reduce the risk of being exposed by model performance detection.

\textbf{ASR:} This metric measures the effectiveness of a one-to-one targeted attack by quantifying the proportion of test samples from the source class that are incorrectly predicted as the designated target class. It is calculated as follows:
\begin{equation}
\text{ASR}=N_{s-to-t}/N_{source}
\end{equation}
Here, $N_{source}$ denotes the total number of test samples belonging to the source class, while $N_{s-to-t}$ represents the subset of those samples that are predicted as the target class. In the PoiCGAN framework, the attack objective is to maximize the ASR value.

\textbf{MIS:} Following the metric design proposed by Xu et al.~\cite{xu2025defense}, we quantify the stealthiness of poisoned models by examining their indistinguishability from benign models in the parameter space.  For instance, in the $m$-th we define the MIS as the inverse of the statistical distance between the benign and poisoned models. Given the high dimensionality of local model parameters, dimensionality reduction is first performed using Principal Component Analysis (PCA)~\cite{fang2023efficient} to project all models into a two-dimensional space. Then, we partition them into two clusters representing the benign and poisoned models, and compute their centroids $u_{m}$ and $u_{m}^{poi}$. The centroids’ calculation is as follows:

\begin{equation}
u_m=\frac{\sum_{i=1}^{N-k} \theta_{m, i}^{(2)}}{N-k} 
\end{equation}
\begin{equation}
u_m^{poi}=\frac{\sum_{j=1}^k \theta_{m, j}^{(2)}}{k}
\end{equation}
where $\theta_{m, i}^{(2)}$ and $\theta_{m, j}^{(2)}$ represent the reduced two-dimensional local models for the benign and poisoned cases, respectively. Prior studies~\cite{jarman2020hierarchical} have shown that greater statistical separation between benign and poisoned models implies higher distinguishability and, consequently, lower stealthiness in the parameter space. Accordingly, we define the MIS as the inverse of the Euclidean distance between $u_{m}$ and $u_{m}^{poi}$, calculated as follows:
\begin{equation}
\text{MIS}=\frac{1}{E(u_{m},u_{m}^{poi})}=\frac{1}{||u_{m}-u_{m}^{poi}||}
\end{equation}
The inverse is taken to visually represent the concealment of the poisoned model; thus, a larger MIS indicates stronger concealment.

\subsection{Poisoning samples visualization}
\label{5.2}
This section provides qualitative evidence of the visual similarity between poisoned samples produced by PoiCGAN and their corresponding clean counterparts. Specifically, poisoned samples generated by the PSG module at the 100th, 200th, 300th, and 400th training rounds across different datasets are illustrated in \tabref{tab4}. As training proceeds, the synthesized images conditioned on the target label progressively converge toward the visual appearance of the source-class samples. This highlights that the PoiCGAN method successfully achieves targeted automatic label flipping during the data generation process.

\begin{table}[h]
    \centering
    \footnotesize 
    \renewcommand{\arraystretch}{1} 
    \caption{Visualization of real samples and generated poisoned samples on various datasets.}
    \label{tab4}
    \setlength{\tabcolsep}{0.65mm} 
    \setlength{\heavyrulewidth}{0.1mm}  
    \setlength{\lightrulewidth}{0.1mm}  
    \begin{tabular}{>{\centering\arraybackslash} m{2cm} m{1.4cm}  m{1.4cm} m{0.4cm} m{1.4cm} m{1.4cm} m{1.4cm} m{1.4cm} }
    \toprule 
    \multirow{2}{*}{\centering  Datasets}  & \multicolumn{2}{c}{Real} &~ & \multicolumn{4}{c}{PoiCGAN, c=t (\textbf{Target})} \\
    ~ &\multicolumn{1}{c}{\textbf{Source}} &\multicolumn{1}{c}{Target}
    &~ &\multicolumn{1}{c}{n=100} &\multicolumn{1}{c}{n=200} &\multicolumn{1}{c}{n=300} &\multicolumn{1}{c}{n=400}\\
    \midrule
     \multicolumn{1}{c}{InsPLAD}  
     &\hspace{0.25 cm}\includegraphics[width=1 cm]{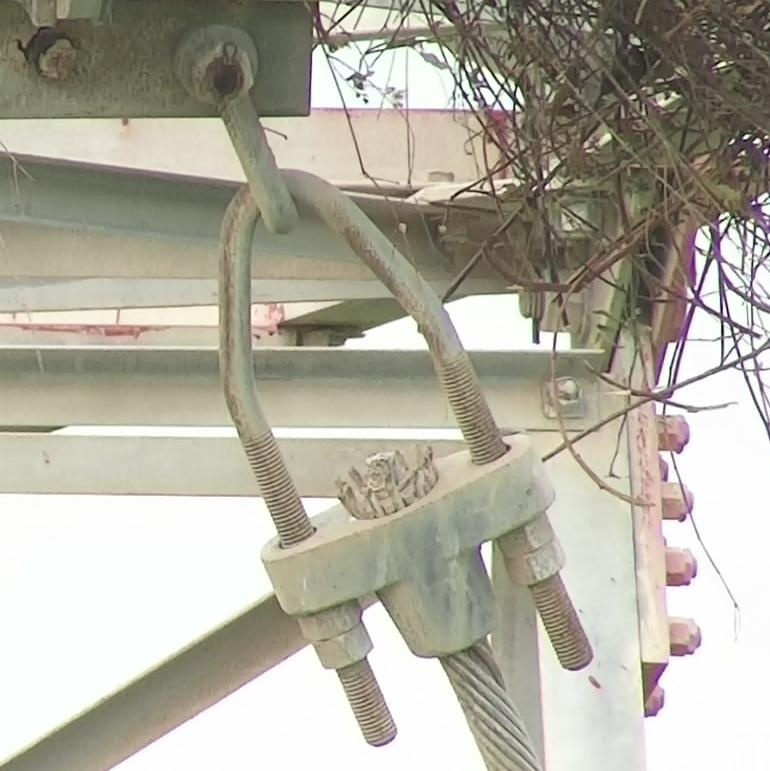} &\hspace{0.25 cm}\includegraphics[width=1 cm]{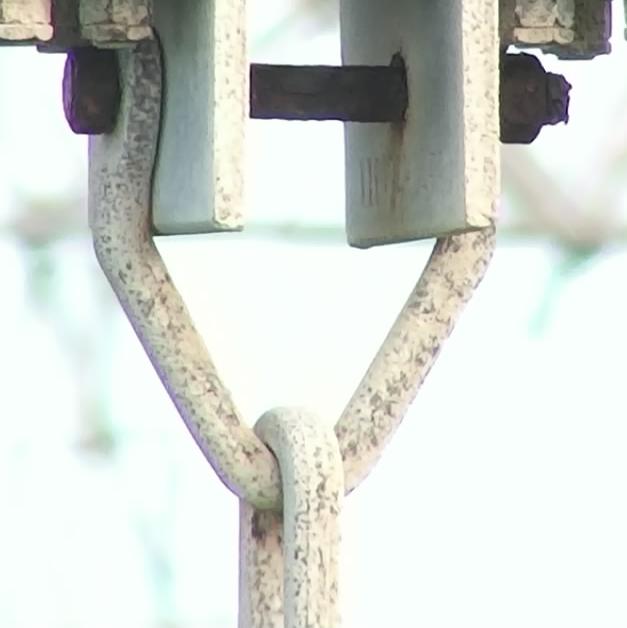}
     &~
     &\hspace{0.25 cm}\includegraphics[width=1 cm]{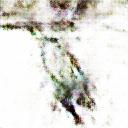} &\hspace{0.25 cm}\includegraphics[width=1 cm]{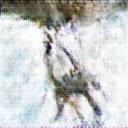} &\hspace{0.2 cm}\includegraphics[width=1 cm]{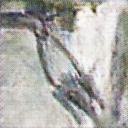} &\hspace{0.2 cm}\includegraphics[width=1 cm]{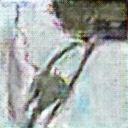} \\ [10pt]
    \multicolumn{1}{c}{NEU-CLS}  
    &\hspace{0.25 cm}\includegraphics[width=1 cm]{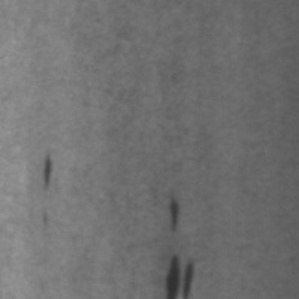} &\hspace{0.25 cm}\includegraphics[width=1 cm]{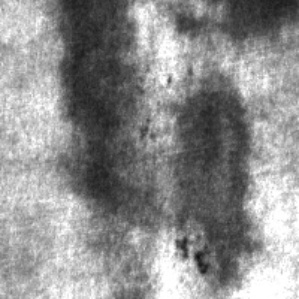} 
    &~
    &\hspace{0.25 cm}\includegraphics[width=1 cm]{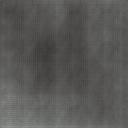} &\hspace{0.25 cm}\includegraphics[width=1 cm]{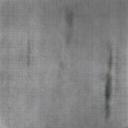} &\hspace{0.2 cm}\includegraphics[width=1 cm]{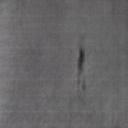} &\hspace{0.2 cm}\includegraphics[width=1 cm]{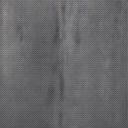} \\ [10pt]
     \multicolumn{1}{c}{Kylberg}  
     &\hspace{0.25 cm}\includegraphics[width=1 cm]{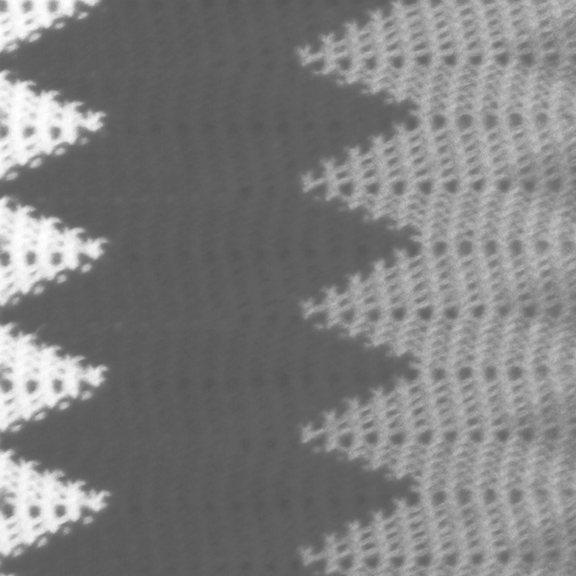} &\hspace{0.25 cm}\includegraphics[width=1 cm]{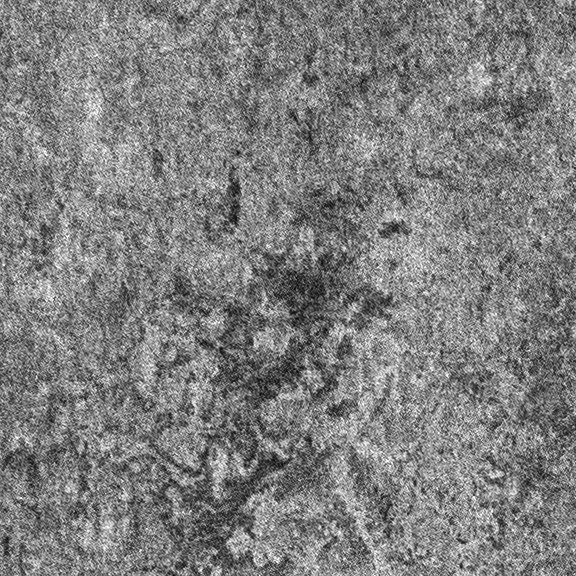} 
     &~
     &\hspace{0.25 cm}\includegraphics[width=1 cm]{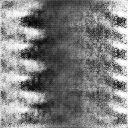} &\hspace{0.25 cm}\includegraphics[width=1 cm]{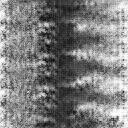} &\hspace{0.2 cm}\includegraphics[width=1 cm]{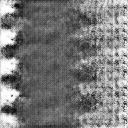} &\hspace{0.2 cm}\includegraphics[width=1 cm]{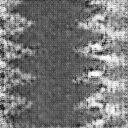} \\ [10pt]
    \bottomrule 
    \end{tabular}
\end{table}

Additionally, we observe that within the same training round, the generated image quality is highest on the InsPLAD dataset and lowest on Kylberg. This discrepancy is due to the greater complexity of the original images on InsPLAD, which results in smaller visual differences. Moreover, owing to the inherent complexity and task-specific characteristics of industrial images~\cite{baitieva2024supervised}, distinguishing poisoned samples from real ones through visual inspection alone—without access to label information—remains challenging, even for experienced observers. This characteristic contributes significantly to PoiCGAN's exceptional concealment ability in industrial image classification scenarios. 

\subsection{Main results}
\label{5.3}
\begin{figure}[h]
\centering
    \subfloat[Attack performance]{\includegraphics[width=0.5\textwidth]{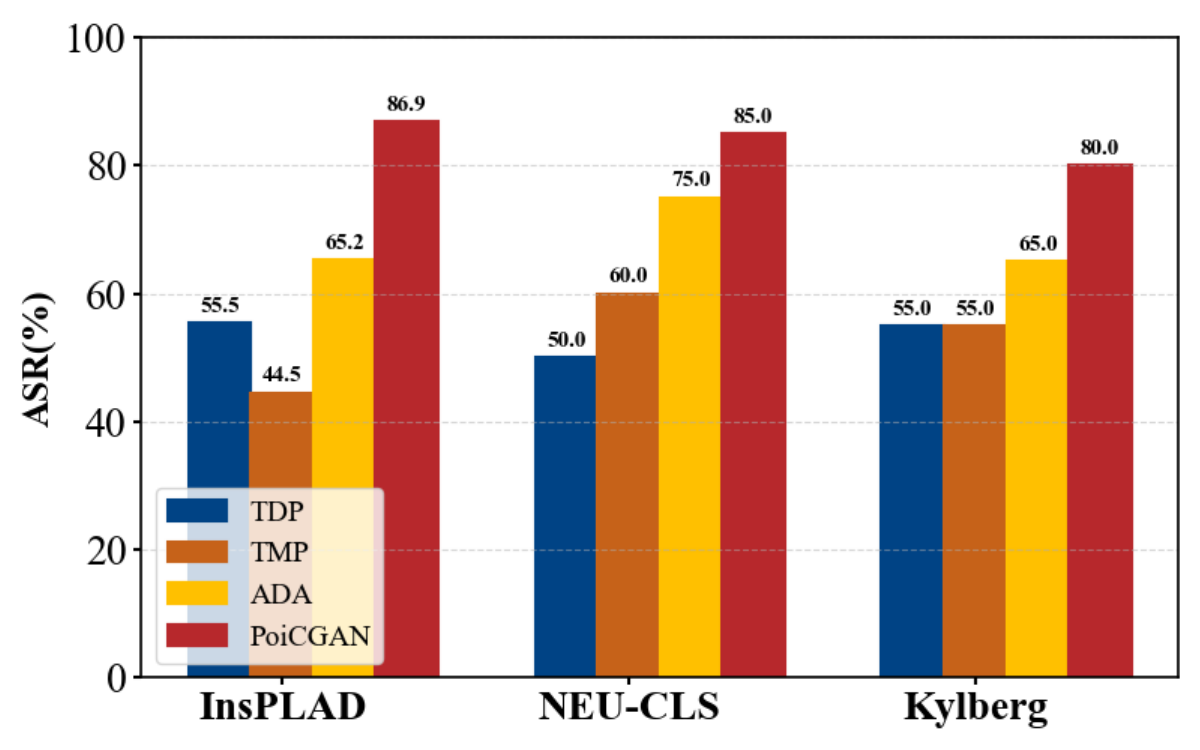}\label{fig2.a}}
    \subfloat[Main task performance]{\includegraphics[width=0.5\textwidth]{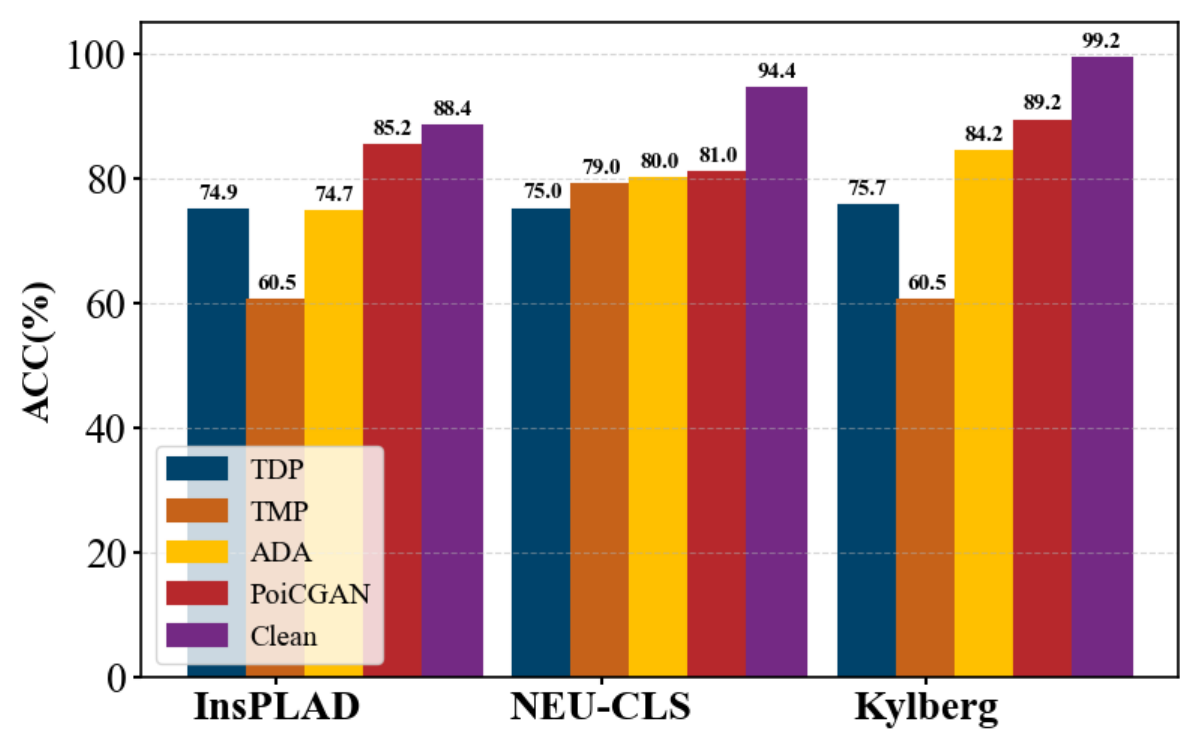}\label{fig2.b}}
    \caption{Comparison of attack performance and main task performance across different methods on various datasets.}
\label{fig2}
\end{figure}

\begin{figure}[h!]
\centering
    \subfloat[TDP]{\includegraphics[width=0.5\textwidth]{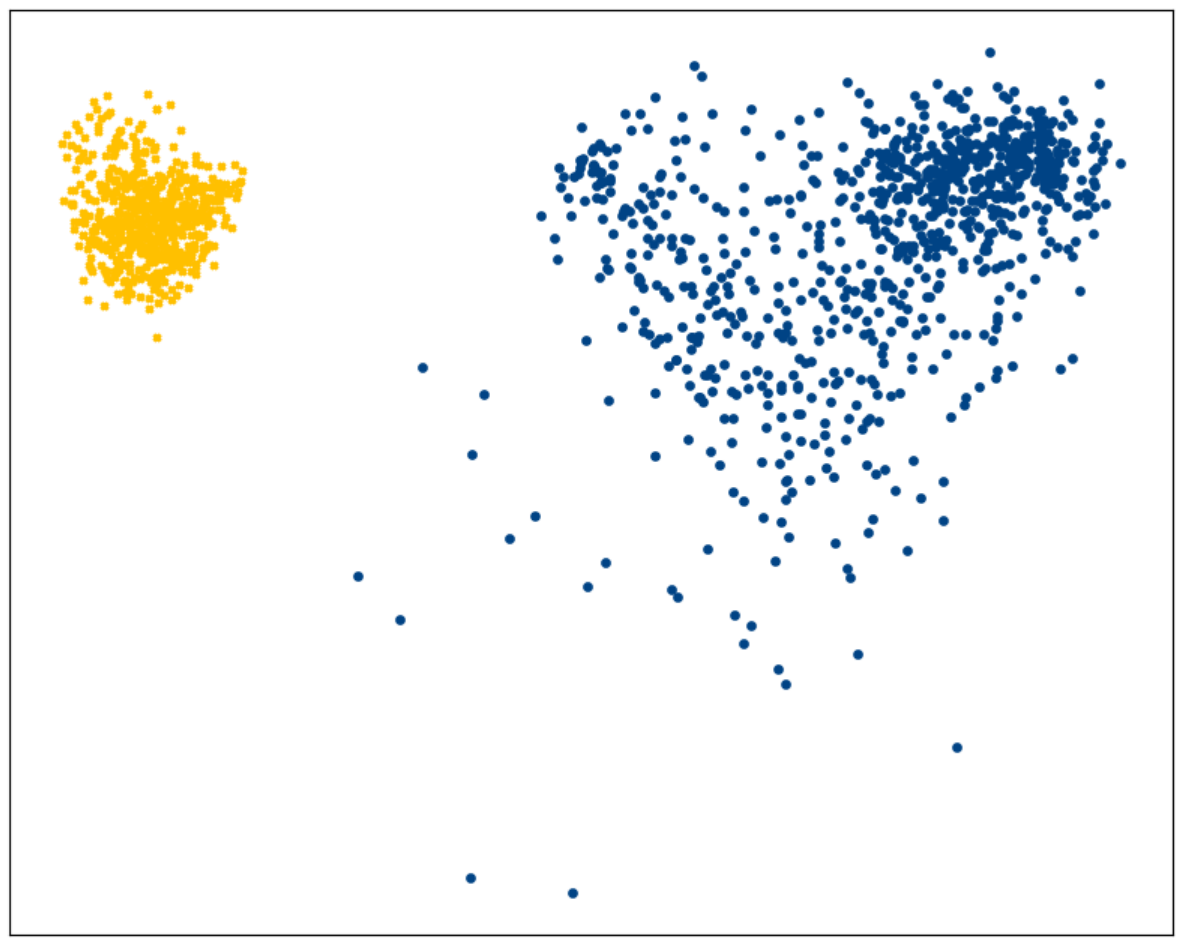}\label{fig3.a}}
    \subfloat[TMP]{\includegraphics[width=0.5\textwidth]{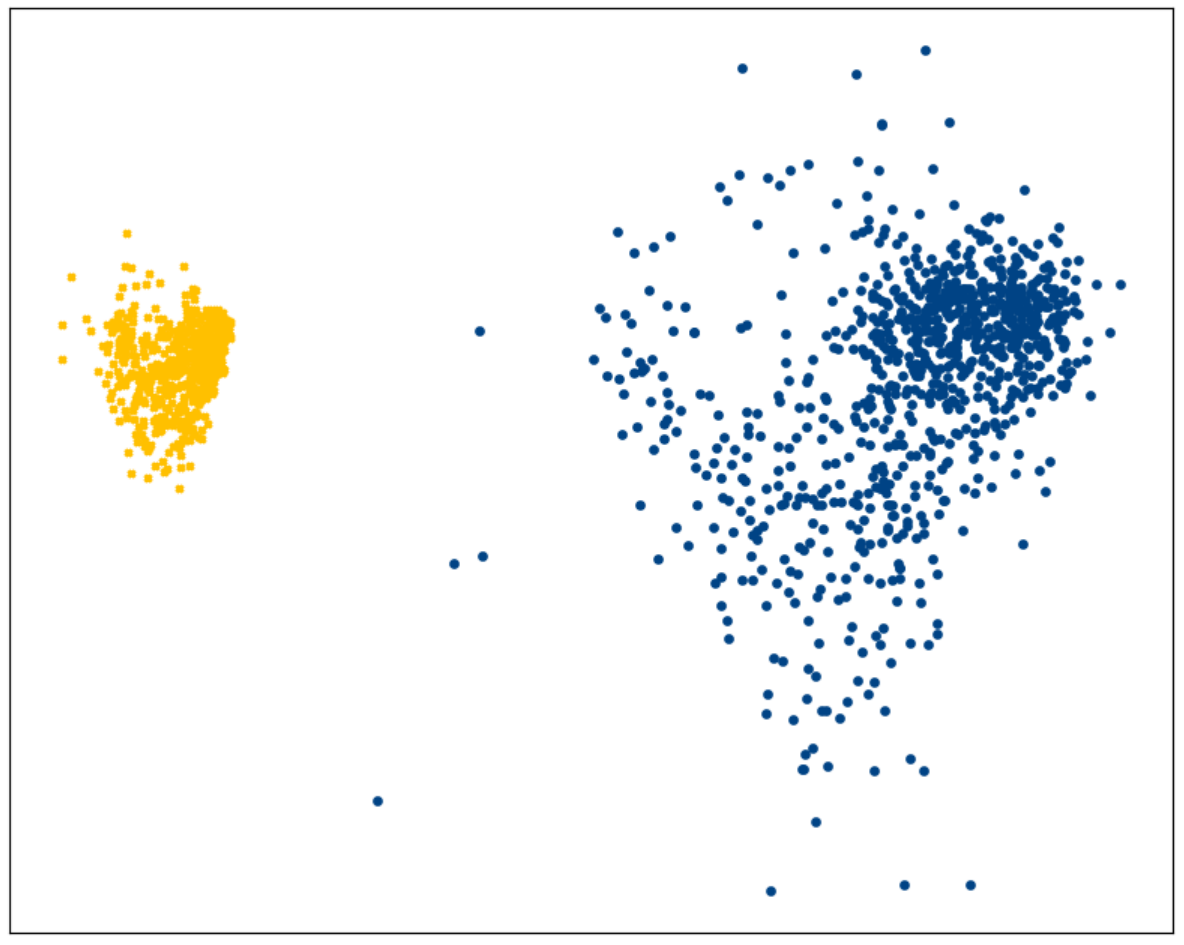}\label{fig3.b}}
    \newline
    \subfloat[ADA]{\includegraphics[width=0.5\textwidth]{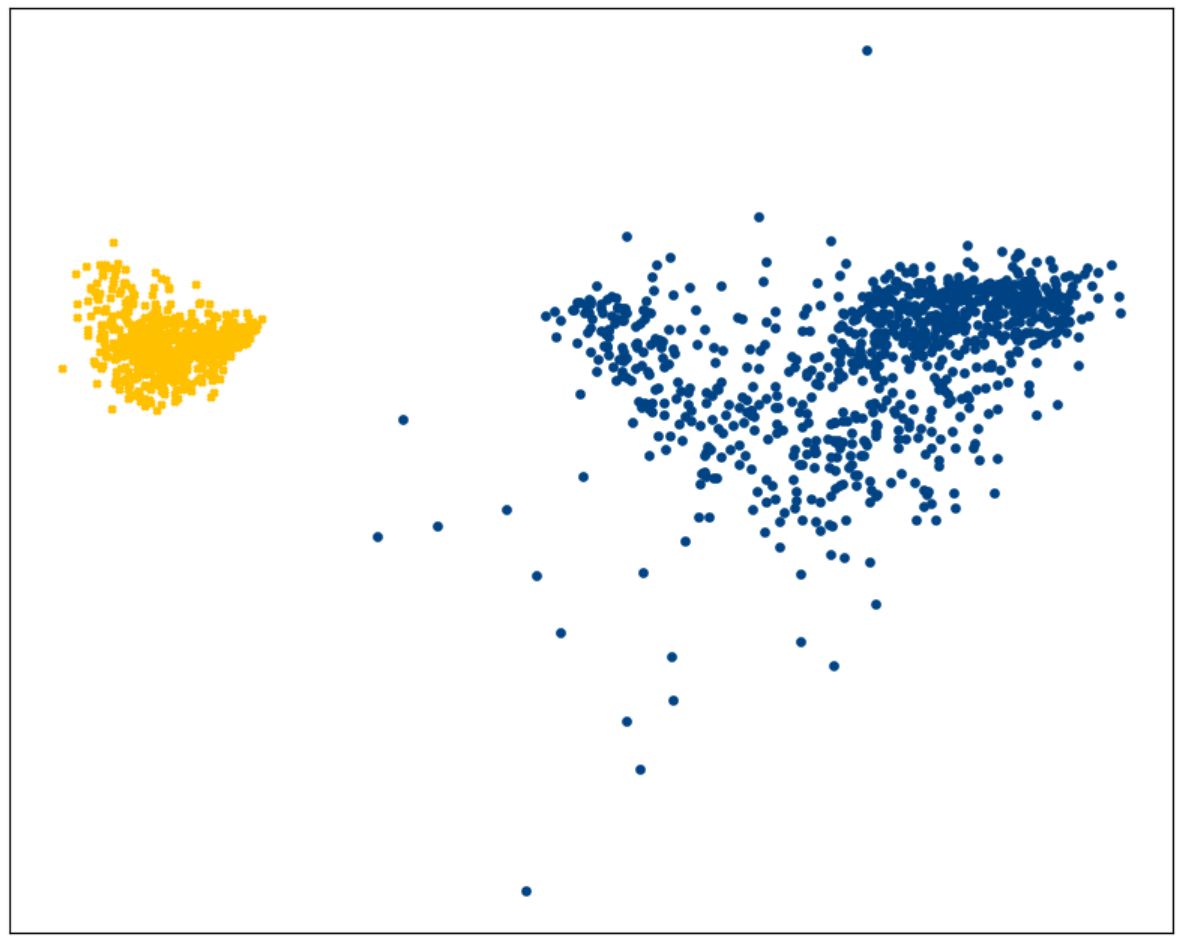}\label{fig3.c}}
    \subfloat[PoiCGAN]{\includegraphics[width=0.5\textwidth]{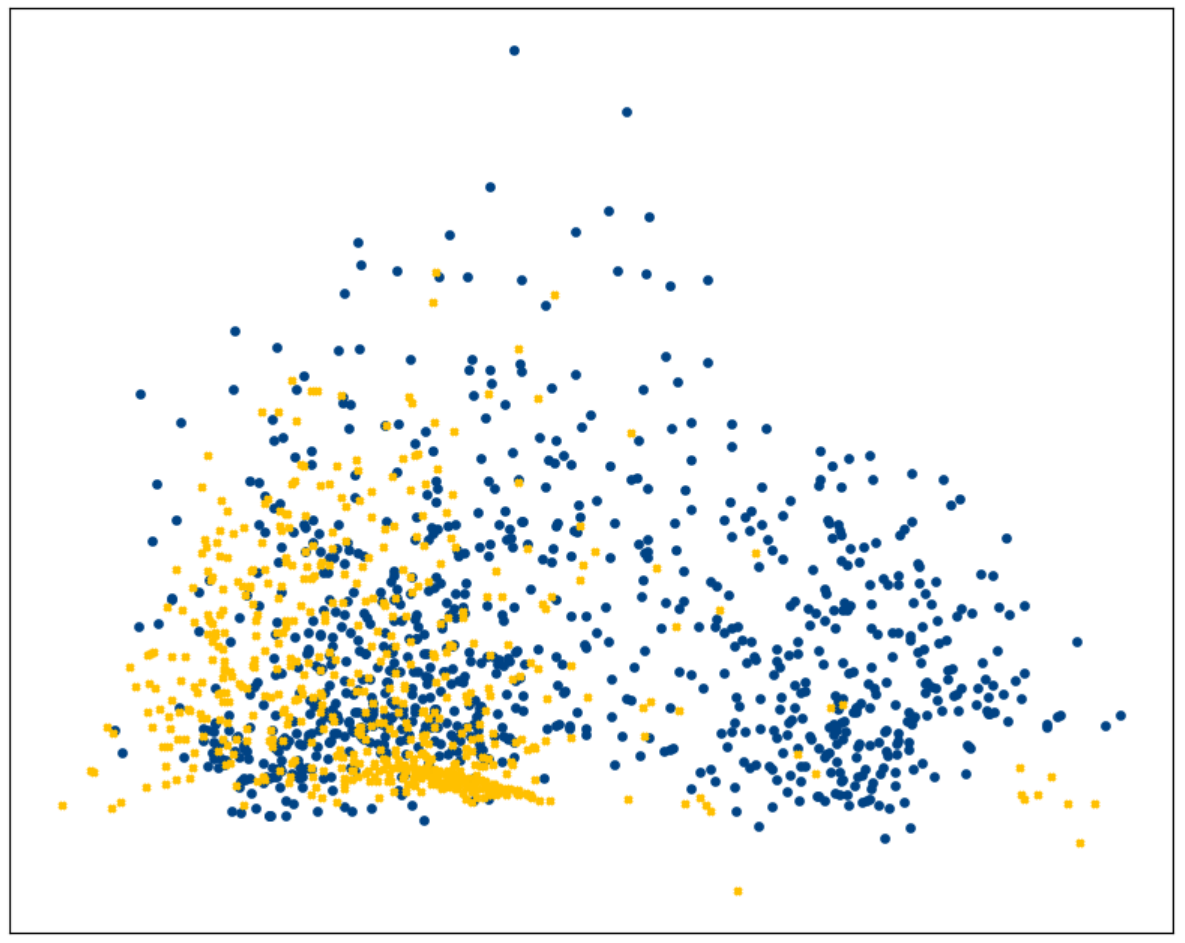}\label{fig3.d}}
    \caption{Visualization of the 2D local model on InsPLAD.}
\label{fig3}
\end{figure}

We validated the attack performance of PoiCGAN and the concealment of the poisoned model on three industrial image classification datasets. We evaluated using the metrics outlined in Section \ref{5.1.4} and compared our method with three advanced baseline approaches to highlight its advantages. \figref{fig2} and \figref{fig3} provide a visual comparison of the attack performance and the concealment of the poisoned model for each method.

\textbf{Attack performance evaluation:}~\figref{fig2} demonstrates that PoiCGAN consistently attains superior ASR and ACC relative to competing baseline methods, indicating the strongest attack performance while maintaining minimal impact on the main task, making it more effective at evading model performance detection. Specifically, our method achieves an average ASR of 83.97\% across the three datasets, surpassing the corresponding values of baseline methods by 15.57\% to 30.8\%, demonstrating the strongest attack performance. This is because PoiCGAN implements automatic label flipping during the generation of poisoned samples. Such attacks, which involve label information influencing the model's learning process, typically have strong signals and clear guidance, resulting in a high ASR.

Moreover, PoiCGAN's average ACC is 85.13\%, showing a decrease of less than 8.9\% compared to when no attack is deployed, while other baseline methods cause an average drop of over 20.1\%, significantly degrading main-task performance. This advantage stems from the adoption of a targeted poisoning strategy, under which prediction deviations are largely confined to the source and target classes. As the number of total main task labels increases, the effect of PoiCGAN on ACC becomes smaller.

\textbf{Concealment of the poisoned model evaluation:} \figref{fig3} visualizes the two-dimensional local models after PCA dimensionality reduction. The yellow and blue points represent the local models of malicious and benign clients, respectively. We observe that the benign and poisoned models form two distinct clusters. For the baseline attacks, the resulting clusters are well separated, with a large distance between their centroids, indicating pronounced discrepancies and limited concealment in the parameter space.  In contrast, for PoiCGAN, the centroids of the two clusters are closer together, and the distributions of the two models are difficult to distinguish, suggesting high concealment of the poisoned model in the model parameter space.

\begin{table}[h]
    \renewcommand{\arraystretch}{0.5} 
    \centering 
    \footnotesize 
    \caption{Comparison of poisoned model stealthiness for different attacks on InsPLAD.}
    \label{tab5}
    \setlength{\tabcolsep}{1.5mm}{
        \setlength{\heavyrulewidth}{0.1mm}  
        \setlength{\lightrulewidth}{0.1mm}  
        \begin{tabular}{c c c c c}
        \toprule 
        ~ &TDP &TMP &ADA &PoiCGAN \\[1.5pt]
        \midrule
        MIS &67.494881 &58.172310 &66.266571 &\textbf{375.224762} \\[1.5pt]
        \bottomrule 
        \end{tabular}}    
\end{table}

We also quantitatively evaluated the poisoned models' concealment using the new metric, MIS, designed in Section \ref{5.1.4}, as shown in \tabref{tab5}. PoiCGAN achieves the highest MIS value, further confirming its excellent model concealment. This behavior can be attributed to the joint perturbation of both feature representations and label assignments during poisoning, which causes the model to internalize poisoned samples in a manner similar to clean data from the target class. As a result, the parameter distributions of poisoned and benign models become increasingly aligned, yielding a reduced statistical separation between them.

\subsection{Hyperparameter sensitivity analyses}
This section discusses three key factors that influence the performance of PoiCGAN: the Poisoned Model Ratio (PMR), the number of training rounds for PSG ($T$), and the scaling factor ($\gamma$). Due to space constraints, only one dataset's results are presented for each experimental setup. This analysis provides insight into how these parameters affect attack behavior and contributes to a more systematic understanding of PoiCGAN’s performance characteristics.

\subsubsection{Impact of poisoned model rate}
\begin{figure}[h]
\centering
\includegraphics[width=0.5\textwidth]{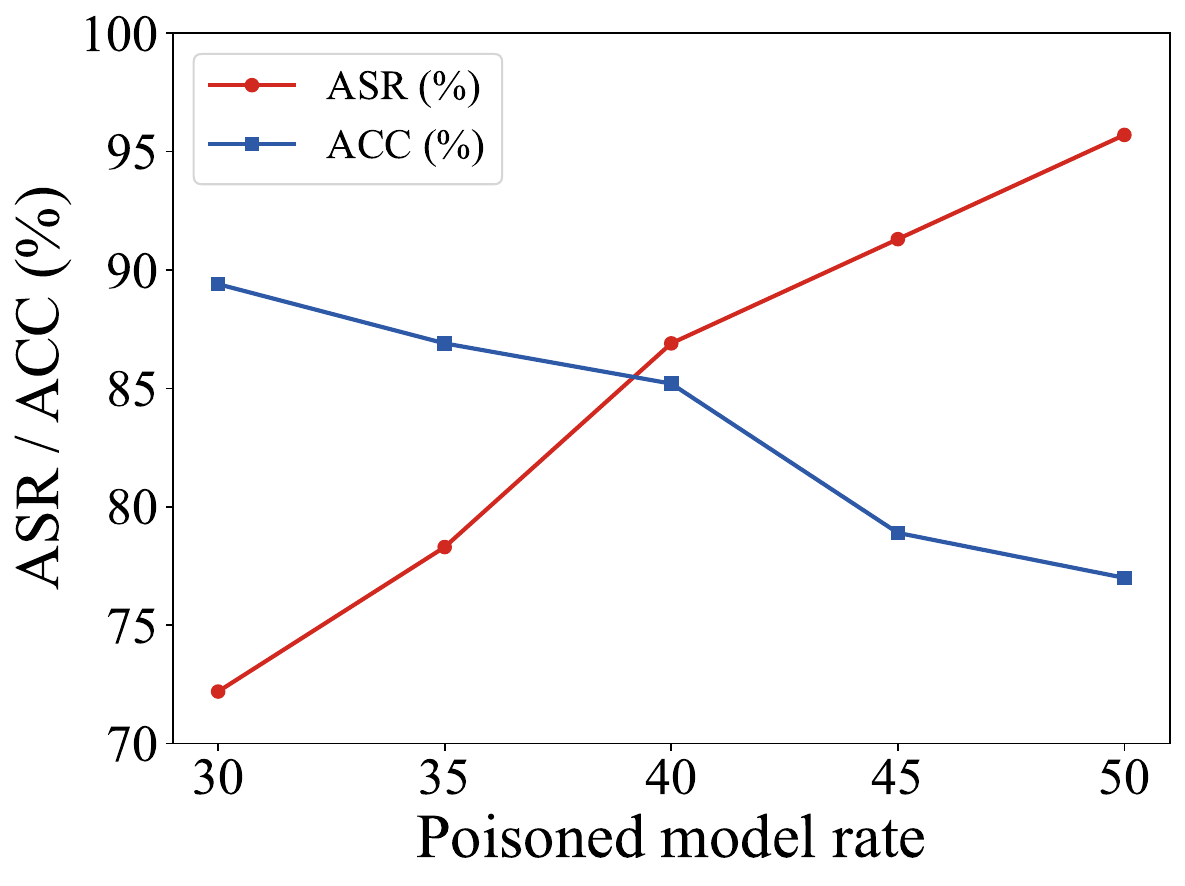}
\caption{Effect of poisoned model rate on attack and main task performance.}
\label{fig4}
\end{figure}

Based on prior research, we hypothesize that the PMR, which represents the proportion of malicious clients among all clients, influences attack performance. To test this, we conducted experiments with different PMR values on InsPLAD. As shown in \figref{fig4}, as the PMR increases, the ASR significantly improves, indicating that PoiCGAN's attack performance strengthens. This behavior arises from the growing influence of poisoned updates during aggregation, which progressively biases the global model toward the adversarial objective and strengthens the targeted attack.

However, the increase in PMR also results in a slight decrease in ACC, as the number of benign models participating in the aggregation decreases, weakening the global model’s responsiveness to the primary task. Empirically, when the PMR is set to 40\%, PoiCGAN maintains a favorable balance between attack effectiveness and task performance.

\subsubsection{Impact of training rounds}
\begin{figure}[h]
\centering
\includegraphics[width=0.5\textwidth]{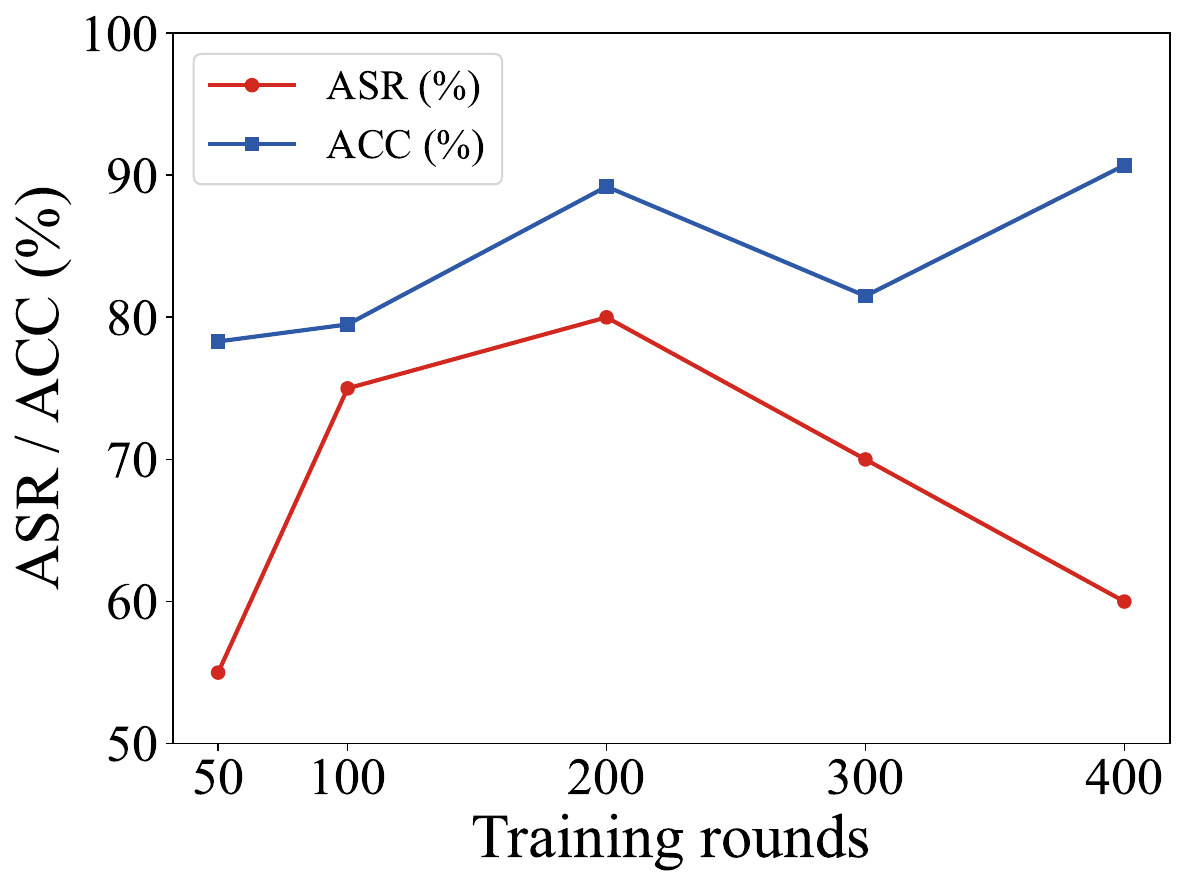}
\caption{Effect of training rounds on attack and main task performance.}
\label{fig5}
\end{figure}

Through extensive experimental validation, we found that the number of training rounds for the PSG module also affects the attack performance. To examine this, we conducted experiments with different $T$ values on Kylberg. As shown in \figref{fig5}, as $T$ increases, ACC shows an overall upward trend. When $T$ is insufficient, the poisoned samples generated by $G_{poi}$ suffer from low visual fidelity, and incorporating such samples into aggregation degrades the global model’s performance on the primary task.

Interestingly, ASR follows a trend of increasing initially, then decreasing as $T$ grows. When the training process is insufficiently iterated, the generated images barely contain any distinct features of the source class, making it difficult for the model to learn these features and establish a clear link~\cite{chen2022linkbreaker} with the target label, resulting in a lower ASR. On the other hand, when the training rounds are too many, the generated images become highly similar to the original source class images, which essentially turns the process into a label-flipping attack. This causes a severe conflict between the targeted attack task and the main task during aggregation, negatively impacting the ASR. Therefore, selecting an optimal number of training rounds is crucial to ensure strong attack performance while minimizing conflicts with the main task. We found that $T=200$ is an ideal choice in this experiment, as it achieves an ASR of 80\% while maintaining an ACC close to 90\%.

\subsubsection{Impact of scaling factor}
\begin{figure}[h]
\centering
\includegraphics[width=0.5\textwidth]{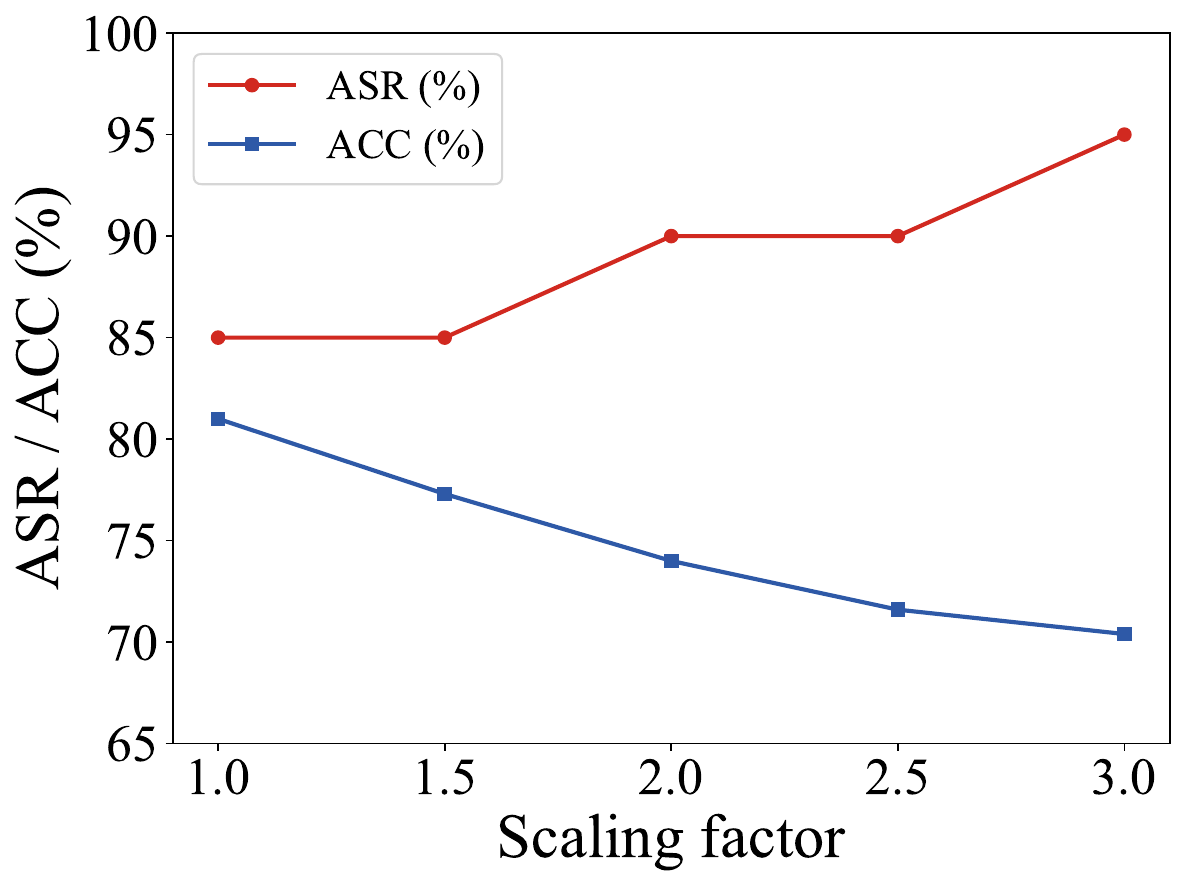}
\caption{Effect of scaling factor on attack and main task performance.}
\label{fig6}
\end{figure}

We also observed that the scaling factor influences the attack performance. To verify this, we conducted experiments with different $\gamma$ values on NEU-CLS. As shown in \figref{fig6}, the ASR significantly improves with the increase of $\gamma$, indicating that PoiCGAN's attack performance strengthens. This trend results from the amplified influence of poisoned updates during aggregation, which biases the global model toward the adversarial objective and strengthens the targeted attack.

However, this increase in the scaling factor leads to a significant drop in ACC, as the contribution of benign models during aggregation decreases, thereby impairing the global model’s performance on the primary task. Therefore, the adversary must balance the performance of both tasks when selecting an appropriate $\gamma$ value. We found that in this experiment, using $\gamma=1$ (i.e., no scaling) resulted in the most balanced overall performance for PoiCGAN. Although further increasing $\gamma$ improves ASR, it does so at the cost of ACC, which is not recommended.

\subsection{Robustness of PoiCGAN against defenses}
\label{5.5}
This section evaluates the robustness of PoiCGAN against three advanced defense methods: Krum, RLR, and FLAME. We begin by briefly describing the principles and features of each defense method. The defenses in this experiment are deployed according to the configurations specified in the original papers.

\textbf{Krum~\cite{blanchard2017machine}:} Krum introduces an aggregation rule based on geometric distance, selecting the gradients closest to the majority of node updates to ensure that distributed training converges to the global optimum, even in the presence of Byzantine nodes. The method is theoretically proven to have the property of ``almost certain convergence”, making it a foundational method for robust distributed training.

\textbf{RLR~\cite{ozdayi2021defending}:} RLR introduces a dynamic learning-rate adaptation strategy that adjusts model updates in response to anomalous client behaviors. By jointly considering both the direction and magnitude of client updates, the method mitigates the influence of malicious contributions on the global model.  This method provides a lightweight, efficient defense without additional computational overhead.

\textbf{FLAME~\cite{nguyen2022flame}:} FLAME is a backdoor defense method that is also fully applicable to poisoning attack defense. It combines dynamic clustering, adaptive pruning, and differential privacy noise injection to create a multi-layer defense system, eliminating attacks while maintaining the performance of benign models. FLAME supports real-time defense in dynamic attack scenarios, balancing both efficiency and effectiveness.

To conduct an entensive evaluation, we deployed the Krum, RLR, and FLAME defense mechanisms on the Kylberg, NEU-CLS, and InsPLAD datasets to assess the performance of PoiCGAN and the baseline attacks. As shown in \tabref{tab6}, almost all of the baseline attack methods failed after the deployment of defenses, with average ASRs of only 16.23\%, 21\%, and 29.57\%. In contrast, PoiCGAN maintained an average ASR of 67.17\%. Even against the recently proposed and highly robust FLAME defense, PoiCGAN achieved an ASR of 56.5\%, which demonstrates the strong resilience of our method to various advanced defense mechanisms. Furthermore, PoiCGAN has minimal impact on the main task performance, remaining at a similar level to that of the other baseline methods.

\begin{table}[h]
    \renewcommand{\arraystretch}{0.5} 
    \centering 
    \footnotesize 
    \caption{Performance of different attack methods under various defense mechanisms.}
    \label{tab6}
    \setlength{\tabcolsep}{0.5mm}{
            \resizebox{0.95\linewidth}{!}{
             \setlength{\heavyrulewidth}{0.1mm}  
            \setlength{\lightrulewidth}{0.1mm}  
            \begin{tabular}{c @{\hspace{2.5mm}} cccc @{\hspace{2.5mm}} cccc @{\hspace{2.5mm}} cccc }
            \toprule 
            Datasets: Defenses  &\multicolumn{4}{c}{InsPLAD: FLAME} &\multicolumn{4}{c}{NEU-CLS: RLR} &\multicolumn{4}{c}{Kylberg: Krum}\\[2.5pt]
            Attacks &TDP &TMP &ADA &PoiCGAN &TDP &TMP &ADA &PoiCGAN &TDP &TMP &ADA &PoiCGAN \\[1.5pt]
            \midrule
            ASR &8.7\% &13.0\% &8.7\% & \textbf{56.5\%} &25.0\% &40.0\% &55.0\% & \textbf{70.0\%} &15.0\% &10.0\% &25.0\% & \textbf{75.0\%} \\[1.5pt]
            ACC &88.6\% &87.8\% &86.2\% & 86.2\% &88.0\% &82.0\% &85.0\% & 77.3\% &86.9\% &91.8\% &86.3\% & 87.8\%\\[1.5pt]
            \bottomrule 
            \end{tabular}}    
        }
\end{table}

\section{Conclusion}
We present PoiCGAN, a stealthy and effective poisoning attack tailored for FL–based industrial image recognition. The key feature of this method is the careful design of the CGAN training process to enable automatic label flipping during the poison sample generation, ensuring attack effectiveness while enhancing the attack's stealthiness. Additionally, our method employs the target label as conditional information for CGAN, facilitating one-to-one targeted attacks and minimizing the impact of poisoning on the main task, thereby evading model performance detection.

In future work, we will explore other extensions of PoiCGAN, such as poisoning attacks targeting text and audio data, as well as poisoning attacks in vertical federated learning and federated transfer learning. We will also investigate related defense strategies.

\section*{CRediT authorship contribution statement}
\textbf{Tao Liu:} Conceptualization, Methodology, Software, Validation, Formal analysis, Investigation, Writing - Original Draft, Visualization. \textbf{Jiguang Lv:} Resources, Writing - Review \& Editing, Project administration, Funding acquisition. \textbf{Dapeng Man:} Conceptualization, Methodology, Resources, Writing - Review \& Editing, Supervision, Project administration, Funding acquisition. \textbf{Weiye Xi:} Formal analysis, Visualization. \textbf{Yaole Li:} Software, Validation. 
\textbf{Feiyu Zhao:} 
Formal analysis, Visualization.
\textbf{Kuiming Wang:} Investigation, Validation.  \textbf{Yingchao Bian:} Investigation, Software. \textbf{Chen Xu:} Supervision, Project administration. \textbf{Wu Yang:} Conceptualization, Resources, Writing - Review \& Editing, Supervision, Project administration, Funding acquisition. 

\section*{Declaration of competing interest}
The authors declare that they have no known competing financial interests or personal relationships that could have appeared to influence the work reported in this paper.

\section*{Acknowledgements}
This work was supported by the Joint Funds of the National Natural Science Foundation of China (No.U21B2019, No.U22A2036), the National Natural Science Foundation of China (No.62272127, No.62406086, No.62572144), and Basic Research Projects of the Central Universities and Colleges (3072025ZN0602).

\section*{Data availability}
Data will be made available on request. The code is available at https:
//github.com/PhD-TaoLiu/PoiCGAN.


\bibliographystyle{elsarticle-num} 
\bibliography{ref}

@article{yang2024slsg,
  title={SLSG: Industrial image anomaly detection with improved feature embeddings and one-class classification},
  author={Yang, Minghui and Liu, Jing and Yang, Zhiwei and Wu, Zhaoyang},
  journal={Pattern Recognition},
  volume={156},
  pages={110862},
  year={2024},
  doi = {10.1016/j.patcog.2024.110862}
}

@inproceedings{chen2024unified,
  title={A unified anomaly synthesis strategy with gradient ascent for industrial anomaly detection and localization},
  author={Chen, Qiyu and Luo, Huiyuan and Lv, Chengkan and Zhang, Zhengtao},
  booktitle={European Conference on Computer Vision},
  pages={37--54},
  year={2024},
  doi = {10.1007/978-3-031-72855-6_3}
}

@article{ettalibi2024ai,
  title={AI and computer vision-based real-time quality control: a review of industrial applications},
  author={Ettalibi, Abdelfatah and Elouadi, Abdelmajid and Mansour, Abdeljebar},
  journal={Procedia Computer Science},
  volume={231},
  pages={212--220},
  year={2024},
  doi = {10.1016/j.procs.2023.12.195}
}

@article{li2025survey,
  title={A survey of deep learning for industrial visual anomaly detection},
  author={Li, Zhuo and Yan, Yuhao and Wang, Xiangheng and Ge, Yifei and Meng, Lin},
  journal={Artificial Intelligence Review},
  volume={58},
  number={9},
  pages={279},
  year={2025},
  doi = {10.1007/s10462-025-11287-7}
}

@article{peng2024industrial,
  title={Industrial surface defect detection and localization using multi-scale information focusing and enhancement GANomaly},
  author={Peng, Jiangji and Shao, Haidong and Xiao, Yiming and Cai, Baoping and Liu, Bin},
  journal={Expert Systems with Applications},
  volume={238},
  pages={122361},
  year={2024},
  doi = {10.1016/j.eswa.2023.122361}
}

@article{chen2023surface,
  title={Surface defect detection of industrial components based on vision},
  author={Chen, Zhendong and Feng, Xuefeng and Liu, Li and Jia, Zhenhong},
  journal={Scientific Reports},
  volume={13},
  number={1},
  pages={22136},
  year={2023},
  doi = {10.1038/s41598-023-49359-9}
}

@article{bai2025comprehensive,
  title={A Comprehensive Survey on Machine Learning Driven Material Defect Detection},
  author={Bai, Jun and Wu, Di and Shelley, Tristan and Schubel, Peter and Twine, David and Russell, John and Zeng, Xuesen and Zhang, Ji},
  journal={ACM Computing Surveys},
  volume={57},
  number={11},
  pages={1--36},
  year={2025},
  doi = {10.1145/3730576}
}

@article{escobar2021quality,
  title={Quality 4.0: a review of big data challenges in manufacturing},
  author={Escobar, Carlos A and McGovern, Megan E and Morales-Menendez, Ruben},
  journal={Journal of Intelligent Manufacturing},
  volume={32},
  number={8},
  pages={2319--2334},
  year={2021},
  doi = {10.1007/s10845-021-01765-4}
}

@article{gao2022review,
  title={A review on recent advances in vision-based defect recognition towards industrial intelligence},
  author={Gao, Yiping and Li, Xinyu and Wang, Xi Vincent and Wang, Lihui and Gao, Liang},
  journal={Journal of manufacturing systems},
  volume={62},
  pages={753--766},
  year={2022},
  doi = {10.1016/j.jmsy.2021.05.008}
}

@article{javaid2022exploring,
  title={Exploring impact and features of machine vision for progressive industry 4.0 culture},
  author={Javaid, Mohd and Haleem, Abid and Singh, Ravi Pratap and Rab, Shanay and Suman, Rajiv},
  journal={Sensors international},
  volume={3},
  pages={100132},
  year={2022},
  doi = {10.1016/j.sintl.2021.100132}
}

@inproceedings{hegiste2024collaborative,
  title={Collaborative learning in shared production environment using federated image classification},
  author={Hegiste, Vinit and Legler, Tatjana and Ruskowski, Martin},
  booktitle={European Symposium on Artificial Intelligence in Manufacturing},
  pages={98--106},
  year={2024},
  doi = {10.1007/978-3-031-86489-6_11}
}

@article{yurdem2024federated,
  title={Federated learning: Overview, strategies, applications, tools and future directions},
  author={Yurdem, Betul and Kuzlu, Murat and Gullu, Mehmet Kemal and Catak, Ferhat Ozgur and Tabassum, Maliha},
  journal={Heliyon},
  volume={10},
  number={19},
  year={2024},
  doi = {10.1016/j.heliyon.2024.e38137}
}

@article{feng2025survey,
  title={A survey of security threats in federated learning},
  author={Feng, Yunhao and Guo, Yanming and Hou, Yinjian and Wu, Yulun and Lao, Mingrui and Yu, Tianyuan and Liu, Gang},
  journal={Complex \& Intelligent Systems},
  volume={11},
  number={2},
  pages={165},
  year={2025},
  doi = {10.1007/s40747-024-01664-0}
}

@article{almutairi2023federated,
  title={Federated learning vulnerabilities, threats and defenses: A systematic review and future directions},
  author={Almutairi, Suzan and Barnawi, Ahmed},
  journal={Internet of Things},
  volume={24},
  pages={100947},
  year={2023},
  doi = {10.1016/j.iot.2023.100947}
}

@article{lu2022defense,
  title={Defense against local model poisoning attacks to byzantine-robust federated learning},
  author={Lu, Shiwei and Li, Ruihu and Chen, Xuan and Ma, Yuena},
  journal={Frontiers of Computer Science},
  volume={16},
  number={6},
  pages={166337},
  year={2022},
  doi = {10.1007/s11704-021-1067-4}
}

@article{moshawrab2024securing,
  title   = {Securing Federated Learning: Approaches, Mechanisms and Opportunities},
  author  = {Moshawrab, Mohammad and Adda, Mehdi and Bouzouane, Abdenour and Ibrahim, Hussein and Raad, Ali},
  journal = {Electronics},
  volume  = {13},
  number  = {18},
  year    = {2024},
  doi = {10.3390/electronics13183675}
}

@article{wang2025enhancing,
  title={Enhancing poisoning attack mitigation in federated learning through perturbation-defense complementarity on history gradients},
  author={Wang, Cong and Mi, Zhilong and Yin, Ziqiao and Guo, Binghui},
  journal={Frontiers of Computer Science},
  volume={19},
  number={12},
  year={2025},
  doi = {10.1007/s11704-025-40924-1}
}

@article{wang2024fedtop,
  title={FedTop: a constraint-loosed federated learning aggregation method against poisoning attack},
  author={Wang, Che and Wu, Zhenhao and Gao, Jianbo and Zhang, Jiashuo and Xia, Junjie and Gao, Feng and Guan, Zhi and Chen, Zhong},
  journal={Frontiers of Computer Science},
  volume={18},
  number={5},
  pages={185348},
  year={2024},
  doi = {10.1007/s11704-024-3767-z}
}

@article{zhao2025data,
  title={Data poisoning in deep learning: A survey},
  author={Zhao, Pinlong and Zhu, Weiyao and Jiao, Pengfei and Gao, Di and Wu, Ou},
  journal={arXiv preprint arXiv:2503.22759},
  year={2025},
  doi = {10.48550/arXiv.2503.22759}
}

@article{yang2024invisible,
  title={Invisible threats in the data: A study on data poisoning attacks in deep generative models},
  author={Yang, Ziying and Zhang, Jie and Wang, Wei and Li, Huan},
  journal={Applied Sciences},
  volume={14},
  number={19},
  pages={8742},
  year={2024},
  doi = {10.3390/app14198742}
}

@article{arazzi2025evading,
  title={Evading model poisoning attacks in federated learning by a long-short-term-memory-based approach},
  author={Arazzi, Marco and Lax, Gianluca and Nocera, Antonino},
  journal={Integrated Computer-Aided Engineering},
  volume={32},
  number={2},
  pages={111--125},
  year={2025},
  doi = {10.1177/10692509241301588}
}

@article{khraisat2025securing,
  title={Securing federated learning: a defense strategy against targeted data poisoning attack},
  author={Khraisat, Ansam and Alazab, Ammar and Alazab, Moutaz and Jan, Tony and Singh, Sarabjot and Uddin, Md Ashraf},
  journal={Discover Internet of Things},
  volume={5},
  number={1},
  pages={16},
  year={2025},
  doi = {10.1007/s43926-025-00108-6}
}

@article{lavaur2025investigating,
  title={Investigating the Impact of Label-flipping Attacks against Federated Learning for Collaborative Intrusion Detection},
  author={Lavaur, L{\'e}o and Busnel, Yann and Autrel, Fabien},
  journal={Computers \& Security},
  volume={156},
  pages={104462},
  year={2025},
  doi = {10.1016/j.cose.2025.104462}
}

@article{ribas2025conditional,
  title={Conditional Generative Adversarial Networks and Deep Learning Data Augmentation: A Multi-Perspective Data-Driven Survey Across Multiple Application Fields and Classification Architectures},
  author={Ribas, Lucas C and Casaca, Wallace and Fares, Ricardo T},
  journal={AI},
  volume={6},
  number={2},
  pages={32},
  year={2025},
  doi = {10.3390/ai6020032}
}

@article{sagar2023poisoning,
  title={Poisoning attacks and defenses in federated learning: A survey},
  author={Sagar, Subhash and Li, Chang-Sun and Loke, Seng W and Choi, Jinho},
  journal={arXiv preprint arXiv:2301.05795},
  year={2023},
  doi = {10.48550/arXiv.2301.05795}
}

@inproceedings{bhagoji2019analyzing,
  title={Analyzing federated learning through an adversarial lens},
  author={Bhagoji, Arjun Nitin and Chakraborty, Supriyo and Mittal, Prateek and Calo, Seraphin},
  booktitle={International conference on machine learning},
  pages={634--643},
  year={2019},
  doi = {10.48550/arXiv.1811.12470}
}

@inproceedings{yin2018byzantine,
  title={Byzantine-robust distributed learning: Towards optimal statistical rates},
  author={Yin, Dong and Chen, Yudong and Kannan, Ramchandran and Bartlett, Peter},
  booktitle={International conference on machine learning},
  pages={5650--5659},
  year={2018},
  doi = {10.48550/arXiv.1803.01498}
}

@article{baruch2019little,
  title={A little is enough: Circumventing defenses for distributed learning},
  author={Baruch, Gilad and Baruch, Moran and Goldberg, Yoav},
  journal={Advances in Neural Information Processing Systems},
  volume={32},
  year={2019},
  doi = {10.48550/arXiv.1902.06156}
}

@article{blanchard2017machine,
  title={Machine learning with adversaries: Byzantine tolerant gradient descent},
  author={Blanchard, Peva and El Mhamdi, El Mahdi and Guerraoui, Rachid and Stainer, Julien},
  journal={Advances in neural information processing systems},
  volume={30},
  year={2017}
}

@inproceedings{shejwalkar2022back,
  title={Back to the drawing board: A critical evaluation of poisoning attacks on production federated learning},
  author={Shejwalkar, Virat and Houmansadr, Amir and Kairouz, Peter and Ramage, Daniel},
  booktitle={2022 IEEE symposium on security and privacy (SP)},
  pages={1354--1371},
  year={2022},
  doi = {10.1109/SP46214.2022.9833647}
}

@inproceedings{fang2020local,
  title={Local model poisoning attacks to $\{$Byzantine-Robust$\}$ federated learning},
  author={Fang, Minghong and Cao, Xiaoyu and Jia, Jinyuan and Gong, Neil},
  booktitle={29th USENIX security symposium (USENIX Security 20)},
  pages={1605--1622},
  year={2020}
}

@article{zhou2021deep,
  title={Deep model poisoning attack on federated learning},
  author={Zhou, Xingchen and Xu, Ming and Wu, Yiming and Zheng, Ning},
  journal={Future Internet},
  volume={13},
  number={3},
  pages={73},
  year={2021},
  doi = {10.3390/fi13030073}
}

@inproceedings{rong2022fedrecattack,
  title={Fedrecattack: Model poisoning attack to federated recommendation},
  author={Rong, Dazhong and Ye, Shuai and Zhao, Ruoyan and Yuen, Hon Ning and Chen, Jianhai and He, Qinming},
  booktitle={2022 IEEE 38th International Conference on Data Engineering (ICDE)},
  pages={2643--2655},
  year={2022},
  doi = {10.1109/ICDE53745.2022.00243}
}

@inproceedings{xie2019dba,
  title={Dba: Distributed backdoor attacks against federated learning},
  author={Xie, Chulin and Huang, Keli and Chen, Pin-Yu and Li, Bo},
  booktitle={International conference on learning representations},
  year={2019}
}

@inproceedings{liu2024beyond,
  title={Beyond traditional threats: A persistent backdoor attack on federated learning},
  author={Liu, Tao and Zhang, Yuhang and Feng, Zhu and Yang, Zhiqin and Xu, Chen and Man, Dapeng and Yang, Wu},
  booktitle={Proceedings of the AAAI Conference on Artificial Intelligence},
  volume={38},
  number={19},
  pages={21359--21367},
  year={2024},
  doi = {10.1609/aaai.v38i19.30131}
}

@article{zhang2020poisongan,
  title={PoisonGAN: Generative poisoning attacks against federated learning in edge computing systems},
  author={Zhang, Jiale and Chen, Bing and Cheng, Xiang and Binh, Huynh Thi Thanh and Yu, Shui},
  journal={IEEE Internet of Things Journal},
  volume={8},
  number={5},
  pages={3310--3322},
  year={2020},
  doi = {10.1109/JIOT.2020.3023126}
}

@inproceedings{sun2023vaguegan,
  title={VagueGAN: a GAN-based data poisoning attack against federated learning systems},
  author={Sun, Wei and Gao, Bo and Xiong, Ke and Lu, Yang and Wang, Yuwei},
  booktitle={2023 20th Annual IEEE International Conference on Sensing, Communication, and Networking (SECON)},
  pages={321--329},
  year={2023},
  doi = {10.1109/SECON58729.2023.10287523}
}

@inproceedings{tolpegin2020data,
  title={Data poisoning attacks against federated learning systems},
  author={Tolpegin, Vale and Truex, Stacey and Gursoy, Mehmet Emre and Liu, Ling},
  booktitle={European symposium on research in computer security},
  pages={480--501},
  year={2020},
  doi = {10.1007/978-3-030-58951-6_24}
}

@inproceedings{sun2022semi,
  title={Semi-targeted model poisoning attack on federated learning via backward error analysis},
  author={Sun, Yuwei and Ochiai, Hideya and Sakuma, Jun},
  booktitle={2022 International Joint Conference on Neural Networks (IJCNN)},
  pages={1--8},
  year={2022},
  doi = {10.1109/IJCNN55064.2022.9891990}
}

@article{gupta2023novel,
  title={A novel data poisoning attack in federated learning based on inverted loss function},
  author={Gupta, Prajjwal and Yadav, Krishna and Gupta, Brij B and Alazab, Mamoun and Gadekallu, Thippa Reddy},
  journal={Computers \& Security},
  volume={130},
  pages={103270},
  year={2023},
  doi = {10.1016/j.cose.2023.103270}
}

@inproceedings{cao2019understanding,
  title={Understanding distributed poisoning attack in federated learning},
  author={Cao, Di and Chang, Shan and Lin, Zhijian and Liu, Guohua and Sun, Donghong},
  booktitle={2019 IEEE 25th international conference on parallel and distributed systems (ICPADS)},
  pages={233--239},
  year={2019},
  doi = {10.1109/ICPADS47876.2019.00042}
}

@article{shi2021federated,
  title={Federated anomaly analytics for local model poisoning attack},
  author={Shi, Siping and Hu, Chuang and Wang, Dan and Zhu, Yifei and Han, Zhu},
  journal={IEEE Journal on Selected Areas in Communications},
  volume={40},
  number={2},
  pages={596--610},
  year={2021},
  doi = {10.1109/JSAC.2021.3118347}
}

@article{cao2020fltrust,
  title={Fltrust: Byzantine-robust federated learning via trust bootstrapping},
  author={Cao, Xiaoyu and Fang, Minghong and Liu, Jia and Gong, Neil Zhenqiang},
  journal={arXiv preprint arXiv:2012.13995},
  year={2020},
  doi = {10.48550/arXiv.2012.13995}
}

@inproceedings{wang2020model,
  title={Model poisoning defense on federated learning: A validation based approach},
  author={Wang, Yuao and Zhu, Tianqing and Chang, Wenhan and Shen, Sheng and Ren, Wei},
  booktitle={International conference on network and system security},
  pages={207--223},
  year={2020},
  doi = {10.1007/978-3-030-65745-1_12}
}

@article{al2023untargeted,
  title={Untargeted poisoning attack detection in federated learning via behavior attestational},
  author={Al Mallah, Ranwa and Lopez, David and Badu-Marfo, Godwin and Farooq, Bilal},
  journal={IEEE Access},
  volume={11},
  pages={125064--125079},
  year={2023},
  doi = {10.1109/ACCESS.2023.3330144}
}

@inproceedings{zhang2022fldetector,
  title={Fldetector: Defending federated learning against model poisoning attacks via detecting malicious clients},
  author={Zhang, Zaixi and Cao, Xiaoyu and Jia, Jinyuan and Gong, Neil Zhenqiang},
  booktitle={Proceedings of the 28th ACM SIGKDD conference on knowledge discovery and data mining},
  pages={2545--2555},
  year={2022},
  doi = {10.1145/3534678.3539231}
}

@inproceedings{mcmahan2017communication,
  title={Communication-efficient learning of deep networks from decentralized data},
  author={McMahan, Brendan and Moore, Eider and Ramage, Daniel and Hampson, Seth and y Arcas, Blaise Aguera},
  booktitle={Artificial intelligence and statistics},
  pages={1273--1282},
  year={2017},
  doi = {10.48550/arXiv.1602.05629}
}

@article{mirza2014conditional,
  title={Conditional generative adversarial nets},
  author={Mirza, Mehdi and Osindero, Simon},
  journal={arXiv preprint arXiv:1411.1784},
  year={2014},
  doi = {10.48550/arXiv.1411.1784}
}

@article{goodfellow2014generative,
  title={Generative adversarial nets},
  author={Goodfellow, Ian J and Pouget-Abadie, Jean and Mirza, Mehdi and Xu, Bing and Warde-Farley, David and Ozair, Sherjil and Courville, Aaron and Bengio, Yoshua},
  journal={Advances in neural information processing systems},
  volume={27},
  year={2014},
  doi = {10.48550/arXiv.1406.2661}
}

@article{he2019semi,
  title={Semi-supervised defect classification of steel surface based on multi-training and generative adversarial network},
  author={He, Yu and Song, Kechen and Dong, Hongwen and Yan, Yunhui},
  journal={Optics and Lasers in Engineering},
  volume={122},
  pages={294--302},
  year={2019},
  doi = {10.1016/j.optlaseng.2019.06.020}
}

@article{macas2024adversarial,
  title={Adversarial examples: A survey of attacks and defenses in deep learning-enabled cybersecurity systems},
  author={Macas, Mayra and Wu, Chunming and Fuertes, Walter},
  journal={Expert Systems with Applications},
  volume={238},
  pages={122223},
  year={2024},
  doi = {10.1016/j.eswa.2023.122223}
}

@article{shannon1949communication,
  title={Communication theory of secrecy systems},
  author={Shannon, Claude E},
  journal={The Bell system technical journal},
  volume={28},
  number={4},
  pages={656--715},
  year={1949},
  doi = {10.1002/j.1538-7305.1949.tb00928.x}
}

@inproceedings{bagdasaryan2020backdoor,
  title={How to backdoor federated learning},
  author={Bagdasaryan, Eugene and Veit, Andreas and Hua, Yiqing and Estrin, Deborah and Shmatikov, Vitaly},
  booktitle={International conference on artificial intelligence and statistics},
  pages={2938--2948},
  year={2020},
  doi = {10.48550/arXiv.1807.00459}
}

@article{yuan2024performance,
  title={Performance analysis of deep learning algorithms implemented using PyTorch in image recognition},
  author={Yuan, Jie},
  journal={Procedia Computer Science},
  volume={247},
  pages={61--69},
  year={2024},
  doi = {10.1016/j.procs.2024.10.008}
}

@article{bouguettaya2023deep,
  title={Deep ensemble transfer learning-based approach for classifying hot-rolled steel strips surface defects},
  author={Bouguettaya, Abdelmalek and Mentouri, Zoheir and Zarzour, Hafed},
  journal={The International Journal of Advanced Manufacturing Technology},
  volume={125},
  number={11},
  pages={5313--5322},
  year={2023},
  doi = {10.1007/s00170-023-10947-8}
}

@article{vieira2023insplad,
  title={Insplad: A dataset and benchmark for power line asset inspection in uav images},
  author={Vieira e Silva, Andr{\'e} Luiz Buarque and de Castro Felix, Heitor and Sim{\~o}es, Franscisco Paulo Magalh{\~a}es and Teichrieb, Veronica and dos Santos, Michel and Santiago, Hemir and Sgotti, Virginia and Lott Neto, Henrique},
  journal={International journal of remote sensing},
  volume={44},
  number={23},
  pages={7294--7320},
  year={2023},
  doi = {10.1080/01431161.2023.2283900}
}

@article{goyal2023texture,
  title={Texture classification for visual data using transfer learning},
  author={Goyal, Vinat and Sharma, Sanjeev},
  journal={Multimedia Tools and Applications},
  volume={82},
  number={16},
  pages={24841--24864},
  year={2023},
  doi = {10.1007/s11042-022-14276-y}
}

@article{an2024lightweight,
  title={A lightweight network architecture for traffic sign recognition based on enhanced LeNet-5 network},
  author={An, Yuan and Yang, Chunyu and Zhang, Shuo},
  journal={Frontiers in neuroscience},
  volume={18},
  pages={1431033},
  year={2024},
  doi = {10.3389/fnins.2024.1431033}
}

@article{nisa2025improved,
  title={Improved Binary Classification of Underwater Images Using a Modified Resnet-18 Model},
  author={Nisa, Mehr and Leszczuk, Mikolaj and Juszka, Dawid and Zhang, Yi},
   journal={Electronics},
  volume={14},
  number={15},
  pages={2954},
  year={2025},
  doi = {10.3390/electronics14152954}
}

@article{shahin2025fine,
  title={Fine-tuned ResNet34 for efficient brain tumor classification},
  author={Shahin, Anas},
  journal={Scientific Reports},
  volume={15},
  number={1},
  pages={36910},
  year={2025},
  doi = {10.1038/s41598-025-20872-3}
}

@article{sun2023attacking,
  title={Attacking-distance-aware attack: Semi-targeted model poisoning on federated learning},
  author={Sun, Yuwei and Ochiai, Hideya and Sakuma, Jun},
  journal={IEEE Transactions on Artificial Intelligence},
  volume={5},
  number={2},
  pages={925--939},
  year={2023},
  doi = {10.1109/TAI.2023.3280155}
}

@article{mcmahan2016federated,
  title={Federated learning of deep networks using model averaging},
  author={McMahan, H Brendan and Moore, Eider and Ramage, Daniel and y Arcas, Blaise Ag{\"u}era},
  journal = {arXiv preprint arXiv:1602.05629},
  year = {2016},
  doi = {10.48550/arXiv.1602.05629}
}

@article{kumar2025minimal,
  title={Minimal data poisoning attack in federated learning for medical image classification: An attacker perspective},
  author={Kumar, K Naveen and Mohan, C Krishna and Cenkeramaddi, Linga Reddy and Awasthi, Navchetan},
  journal={Artificial Intelligence in Medicine},
  volume={159},
  pages={103024},
  year={2025},
  doi = {10.1016/j.artmed.2024.103024}
}

@article{xu2025defense,
  title={A defense mechanism for federated learning in AIoT through critical gradient dimension extraction},
  author={Xu, Jian and Guo, Bing and Chen, Fei and Shen, Yan and Dai, Shengxin and Dai, Cheng and Hu, Yuchuan},
  journal={Computer Communications},
  volume={236},
  pages={108114},
  year={2025},
  doi = {10.1016/j.comcom.2025.108114}
}

@inproceedings{fang2023efficient,
  title={Efficient robust principal component analysis via block Krylov iteration and cur decomposition},
  author={Fang, Shun and Xu, Zhengqin and Wu, Shiqian and Xie, Shoulie},
  booktitle={Proceedings of the IEEE/CVF conference on computer vision and pattern recognition},
  pages={1348--1357},
  year={2023},
  doi = {10.1109/CVPR52729.2023.00136}
}

@article{jarman2020hierarchical,
  title={Hierarchical cluster analysis: Comparison of single linkage, complete linkage, average linkage and centroid linkage method},
  author={Jarman, Angur Mahmud},
  journal={Georgia Southern University},
  volume={29},
  pages={90240},
  year={2020}
}

@inproceedings{baitieva2024supervised,
  title={Supervised anomaly detection for complex industrial images},
  author={Baitieva, Aimira and Hurych, David and Besnier, Victor and Bernard, Olivier},
  booktitle={Proceedings of the IEEE/CVF Conference on Computer Vision and Pattern Recognition},
  pages={17754--17762},
  year={2024},
  doi = {10.1109/CVPR52733.2024.01681}
}

@inproceedings{ozdayi2021defending,
  title={Defending against backdoors in federated learning with robust learning rate},
  author={Ozdayi, Mustafa Safa and Kantarcioglu, Murat and Gel, Yulia R},
  booktitle={Proceedings of the AAAI conference on artificial intelligence},
  volume={35},
  number={10},
  pages={9268--9276},
  year={2021},
  doi = {10.1609/aaai.v35i10.17118}
}

@inproceedings{nguyen2022flame,
  title={$\{$FLAME$\}$: Taming backdoors in federated learning},
  author={Nguyen, Thien Duc and Rieger, Phillip and Chen, Huili and Yalame, Hossein and M{\"o}llering, Helen and Fereidooni, Hossein and Marchal, Samuel and Miettinen, Markus and Mirhoseini, Azalia and Zeitouni, Shaza and others},
  booktitle={31st USENIX Security Symposium (USENIX Security 22)},
  pages={1415--1432},
  year={2022}
}

@article{chen2022linkbreaker,
  title={LinkBreaker: Breaking the backdoor-trigger link in DNNs via neurons consistency check},
  author={Chen, Zhenzhu and Wang, Shang and Fu, Anmin and Gao, Yansong and Yu, Shui and Deng, Robert H},
  journal={IEEE Transactions on Information Forensics and Security},
  volume={17},
  pages={2000--2014},
  year={2022},
  doi = {10.1109/TIFS.2022.3175616}
}





\end{document}